\documentclass[11pt,headsepline,a4paper,chapterinoneline]{report} 
 \pdfoutput=1
\usepackage[utf8]{inputenc}
\usepackage[Sonny]{fncychap}

\providecommand{\keywords}[1]{\textbf{\textit{Keywords---}} #1}

\usepackage[margin=3.0cm]{geometry}
\usepackage{fancyhdr}
\usepackage{palatino}
\usepackage[nomain,acronym,xindy,toc,nonumberlist]{glossaries} 
\makeglossaries
\usepackage[xindy]{imakeidx}
\makeindex

\newenvironment{dedication}
    {\vspace{6ex}\begin{quotation}\begin{center}\begin{em}}
    {\par\end{em}\end{center}\end{quotation}}

\usepackage{graphicx}
\usepackage[pdftex,
            pdfauthor={Sina Ahmadi},
            pdftitle={Attention-Based Encoder-Decoder Networks for Spelling and Grammatical Error Correction},
            pdfsubject={master's thesis},
            pdfkeywords={Natural language error correction, recurrent neural networks, encoder-decoder models, attention mechanism},
            pdfproducer={Latex},
            pdfcreator={PDFLatex}]{hyperref}
\usepackage{float}
\usepackage{wrapfig}
\usepackage[document]{ragged2e}
\graphicspath{ {Pictures/} }
\graphicspath{{imgs/}}
\usepackage{tikz}
\tikzset{basic/.style={draw,fill=blue!50!green!20,
                       text badly centered,minimum width=3em}}
\tikzset{input/.style={basic,circle}}
\tikzset{weights/.style={basic,rectangle,minimum width=2em}}
\tikzset{functions/.style={basic,circle,fill=blue!50!green!20}}
         
\usepackage{pgfplots}
\usetikzlibrary{decorations.pathreplacing}\usetikzlibrary{fadings}

\definecolor{rosso}{RGB}{220,57,18}
\definecolor{giallo}{RGB}{255,153,0}
\definecolor{blu}{RGB}{102,140,217}
\definecolor{verde}{RGB}{16,150,24}
\definecolor{viola}{RGB}{153,0,153}

\makeatletter

\tikzstyle{chart}=[
    legend label/.style={font={\scriptsize},anchor=west,align=left},
    legend box/.style={rectangle, draw, minimum size=5pt},
    axis/.style={black,semithick,->},
    axis label/.style={anchor=east,font={\tiny}},
]
\tikzstyle{bar chart}=[
    chart,
    bar width/.code={
        \pgfmathparse{##1/2}
        \global\let\bar@w\pgfmathresult
    },
    bar/.style={very thick, draw=white},
    bar label/.style={font={\bf\small},anchor=north},
    bar value/.style={font={\footnotesize}},
    bar width=.75,
]
\tikzstyle{pie chart}=[
    chart,
    slice/.style={line cap=round, line join=round, very thick,draw=white},
    pie title/.style={font={\bf}},
    slice type/.style 2 args={
        ##1/.style={fill=##2},
        values of ##1/.style={}
    }
]
\pgfdeclarelayer{background}
\pgfdeclarelayer{foreground}
\pgfsetlayers{background,main,foreground}

\usepackage{multirow}
\newcommand{\pie}[3][]{
    \begin{scope}[#1]
    \pgfmathsetmacro{\curA}{90}
    \pgfmathsetmacro{\r}{1}
    \def\c{(0,0)}
    \node[pie title] at (90:1.3) {#2};
    \foreach \v/\s in{#3}{
        \pgfmathsetmacro{\deltaA}{\v/100*360}
        \pgfmathsetmacro{\nextA}{\curA + \deltaA}
        \pgfmathsetmacro{\midA}{(\curA+\nextA)/2}

        \path[slice,\s] \c
            -- +(\curA:\r)
            arc (\curA:\nextA:\r)
            -- cycle;
        \pgfmathsetmacro{\d}{max((\deltaA * -(.5/50) + 1) , .5)}

        \begin{pgfonlayer}{foreground}
        \path \c -- node[pos=\d,pie values,values of \s]{$\v\%$} +(\midA:\r);
        \end{pgfonlayer}

        \global\let\curA\nextA
    }
    \end{scope}
}
\newcommand{\legend}[2][]{
    \begin{scope}[#1]
    \path
        \foreach \n/\s in {#2}
            {
                  ++(0,-10pt) node[\s,legend box] {} +(5pt,0) node[legend label] {\n}
            }
    ;
    \end{scope}
}
\usetikzlibrary{fit,positioning,arrows.meta,calc,shapes}
\tikzset{neuron/.style={shape=circle, minimum size=1.25cm, 
  inner sep=0, draw, font=\small}, io/.style={neuron, fill=gray!20}}

\tikzset{
  redondo/.style={
    draw=blue,
    line width=1pt,
    rounded corners=3pt,
    text width=#1
  },
  punto/.style={
    fill=red,
    circle,
    inner sep=1.25pt
  },
  tresp/.pic={
    \node[punto] at (0.25,0) {};
    \node[punto] at (0.5,0) {};
    \node[punto] at (0.75,0) {};
  },
  dosp/.pic={
    \node[punto] at (0.25,0) {};
    \node[punto] at (0.5,0) {};
  },
  cuadra/.style={
    fill=teal,
    minimum size=10pt
  },
  arr/.style={
    line width=1pt,
    draw=green!70!black,
    ->,
    >=latex
  }  
}
  
\usepackage{array}
\usepackage{enumitem}
\usepackage{listings} 
\usepackage{color}

\definecolor{codegreen}{rgb}{0,0.6,0}
\definecolor{codegray}{rgb}{0.5,0.5,0.5}
\definecolor{codepurple}{rgb}{0.58,0,0.82}
\definecolor{backcolour}{rgb}{0.95,0.95,0.92}
 
\lstdefinestyle{mystyle}{
    backgroundcolor=\color{backcolour},   
    commentstyle=\color{codegreen},
    keywordstyle=\color{magenta},
    numberstyle=\tiny\color{codegray},
    stringstyle=\color{codepurple},
    basicstyle=\footnotesize,
    breakatwhitespace=false,         
    breaklines=true,                 
    captionpos=b,                    
    keepspaces=true,                 
    numbers=left,                    
    numbersep=5pt,                  
    showspaces=false,                
    showstringspaces=false,
    showtabs=false,                  
    tabsize=2
}
 
\newcounter{example}[section]

\usepackage{amsmath}

\usepackage{pict2e,picture}
\newsavebox\CBox
\newlength\CLength
\def\Circled#1{\sbox\CBox{#1}%
  \ifdim\wd\CBox>\ht\CBox \CLength=\wd\CBox\else\CLength=\ht\CBox\fi
    \makebox[1.2\CLength]{\makebox(0,1.2\CLength){\put(0,0){\circle{1.2\CLength}}}%
    \makebox(0,1.2\CLength){\put(-.5\wd\CBox,0){#1}}}}

\definecolor {processblue}{cmyk}{0.96,0,0,0}
\usepackage{tkz-graph}

\usepackage{multicol}

\definecolor{gray}{rgb}{0.4,0.4,0.4}
\definecolor{darkblue}{rgb}{0.0,0.0,0.6}
\definecolor{cyan}{rgb}{0.0,0.6,0.6}

\lstdefinelanguage{XML}
{
  morestring=[b]",
  morestring=[s]{>}{<},
  morecomment=[s]{<?}{?>},
  morecomment=[s][\color{orange}]{<!--}{-->},
  stringstyle=\color{black},
  identifierstyle=\color{darkblue},
  keywordstyle=\color{cyan},
  morekeywords={xmlns,version,type}
}

\lstdefinestyle{listXML}{language=XML, extendedchars=true,  belowcaptionskip=5pt, xleftmargin=1.8em, xrightmargin=0.5em, numbers=left, numberstyle=\small\ttfamily\bf, frame=single, breaklines=true, breakatwhitespace=true, breakindent=0pt, emph={}, emphstyle=\color{red}, basicstyle=\small\ttfamily, columns=fullflexible, showstringspaces=false, commentstyle=\color{gray}\upshape, linewidth=10cm}

\newsavebox{\myXMLbox}

\hypersetup{
  colorlinks   = true, 
  urlcolor     = blue, 
  linkcolor    = blue, 
  citecolor   = red
}

\usepackage[most]{tcolorbox}

\usepackage{xparse}
\usepackage{lipsum}

\def\exampletext{Example} 

\NewDocumentEnvironment{testexample}{ O{} }
{
\colorlet{colexam}{red!55!black} 
\newtcolorbox[use counter=testexample]{testexamplebox}{%
    empty,
    title={\exampletext: #1},
    attach boxed title to top left,
       minipage boxed title,
    boxed title style={empty,size=minimal,toprule=0pt,top=4pt,left=3mm,overlay={}},
    coltitle=colexam,fonttitle=\bfseries,
    before=\par\medskip\noindent,parbox=false,boxsep=0pt,left=3mm,right=0mm,top=2pt,breakable,pad at break=0mm,
       before upper=\csname @totalleftmargin\endcsname0pt, 
    overlay unbroken={\draw[colexam,line width=.5pt] ([xshift=-0pt]title.north west) -- ([xshift=-0pt]frame.south west); },
    overlay first={\draw[colexam,line width=.5pt] ([xshift=-0pt]title.north west) -- ([xshift=-0pt]frame.south west); },
    overlay middle={\draw[colexam,line width=.5pt] ([xshift=-0pt]frame.north west) -- ([xshift=-0pt]frame.south west); },
    overlay last={\draw[colexam,line width=.5pt] ([xshift=-0pt]frame.north west) -- ([xshift=-0pt]frame.south west); },%
    }
\begin{testexamplebox}}
{\end{testexamplebox}\endlist}

\tikzset{%
  every neuron/.style={
    circle,
    draw,
    minimum size=7mm
  },
  neuron missing/.style={
    draw=none, 
    scale=4,
    text height=0.333cm,
    execute at begin node=\color{black}$\vdots$
  },
}
\usepackage{rotating}

\tikzset{basic/.style={draw,fill=green!50,text width=1em,text badly centered}}
\tikzset{input/.style={basic,circle}}
\tikzset{weights/.style={basic,rectangle,fill=green!10}}
\tikzset{functions/.style={basic,circle,fill=blue!10}}

\usepackage[nodisplayskipstretch]{setspace}
\usepackage{amssymb}

\newenvironment{customlegend}[1][]{%
    \begingroup
    \csname pgfplots@init@cleared@structures\endcsname
    \pgfplotsset{#1}%
}{%
    \csname pgfplots@createlegend\endcsname
    \endgroup
}%

\def\addlegendimage{\csname pgfplots@addlegendimage\endcsname}

\pgfkeys{/pgfplots/number in legend/.style={%
        /pgfplots/legend image code/.code={%
            \node at (0.125,-0.0225){#1}; 
        },%
    },
}
\pgfplotsset{
every legend to name picture/.style={west}
}

\usepackage{datetime}

\newdateformat{monthyeardate}{%
  \monthname[\THEMONTH], \THEYEAR}
  
\usepackage{algorithm}
\usepackage[noend]{algpseudocode}

\makeatletter
\def\BState{\State\hskip-\ALG@thistlm}
\makeatother
\makeglossaries
\begin{document}

\begin{titlepage}
\newcommand{\HRule}{\rule{\linewidth}{0.5mm}} 
\center 
\includegraphics[scale=.09]{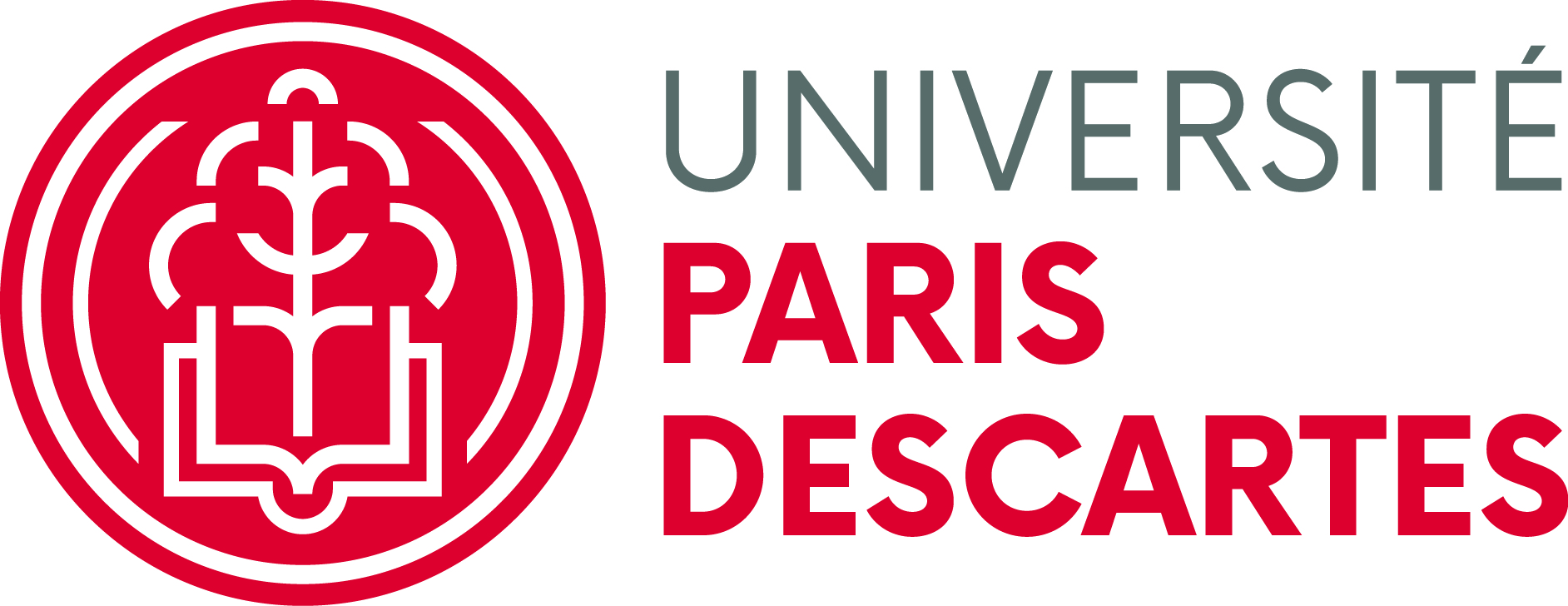} 
\hfill
\includegraphics[scale=.06]{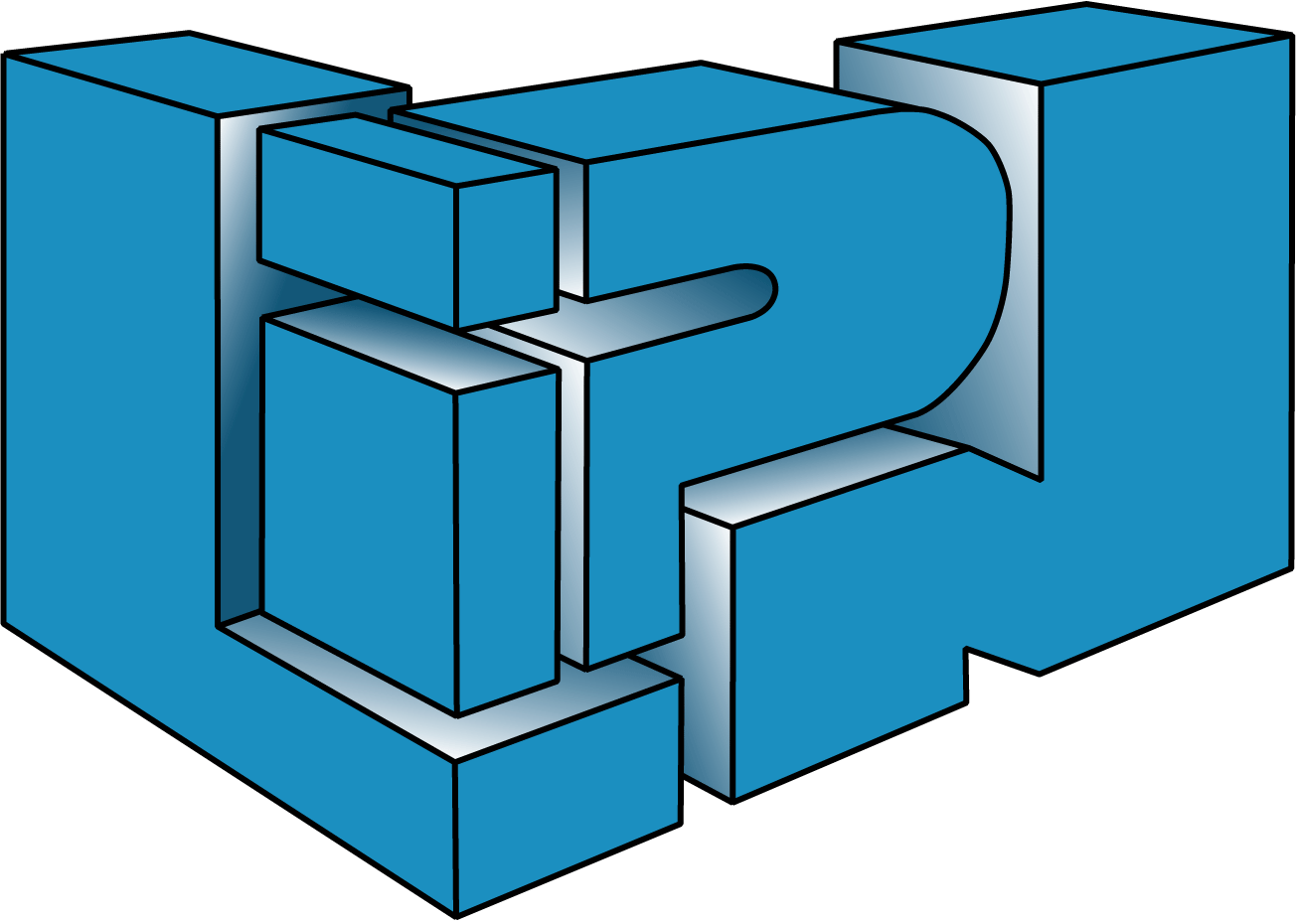}\\[2cm] 

\textsc{\LARGE Paris Descartes University}\\ 
\textsc{\Large Department of Mathematics and Computer Science}\\[0.5cm] 
\textsc{\large Master's thesis}\\[0.5cm] 
\HRule \\[0.5cm]
 \begin{LARGE}
\bfseries Attention-based Encoder-Decoder Networks for Spelling and Grammatical Error Correction
\end{LARGE} \\[0.4cm] 
\HRule \\[1cm]

\begin{minipage}{0.4\textwidth}
\begin{flushleft} \large
\emph{Author:}\\
\href{http://sinaahmadi.github.io/}{Sina \textsc{Ahmadi} }
\end{flushleft}
\end{minipage}
~
\begin{minipage}{0.4\textwidth}
\begin{flushright} \large
\emph{Supervisors:} \\
\href{http://lipn.univ-paris13.fr/~leroux/}{Joseph \textsc{Le Roux}} \\
\href{http://lipn.univ-paris13.fr/~tomeh/}{Nadi \textsc{Tomeh}}
\end{flushright}
\end{minipage}\\[5cm]

\large \textit{A thesis submitted in partial fulfillment of the requirements for the degree of\\ Master of Science in Artificial Intelligence}\\[2cm] 
{\large September, 2017}\\[2cm] 
\end{titlepage}

\justifying

\renewcommand{\thepage}{\roman{page}}
\begin{dedication}
Dedicated to Dad and Mom,\\
and my love, Ioanna
\end{dedication}
\clearpage
    
\chapter*{Acknowledgements}
\addcontentsline{toc}{chapter}{Acknowledgements}

I would first like to thank my thesis supervisors Dr. Joseph Le Roux and Dr. Nadi Tomeh of RCLN team at the Laboratoire d'Informatique de Paris Nord at the University of Paris 13. During the six months of my internship, they were always around whenever I ran into a trouble spot or had a question about my research or my codes. They consistently allowed this thesis to be my own work, but steered me in the right direction.

I would also like to acknowledge the head of the Machine Learning for Data Science (MLDS) team at Paris Descartes University, Dr. Mohamed Nadif. Furthermore, I am gratefully indebted to Dr. Kyumars S. Esmaili and Jalal Sajadi for their very valuable presence and friendship during my studies.

Finally, I must express my very profound gratitude to my parents, my wife Ioanna and our whole family, for their unconditional love, their forbearance and for providing me with unfailing support and continuous encouragement throughout my years of study and through the process of researching and writing this thesis. This accomplishment would not have been possible without them.

\clearpage

\chapter*{Abstract}
\addcontentsline{toc}{chapter}{Abstract}

Automatic spelling and grammatical correction systems are one of the most widely used tools within natural language applications. In this thesis, we assume the task of error correction as a type of monolingual machine translation where the source sentence is potentially erroneous and the target sentence should be the corrected form of the input. Our main focus in this project is building neural network models for the task of error correction. In particular, we investigate sequence-to-sequence and attention-based models which have recently shown a higher performance than the state-of-the-art of many language processing problems. We demonstrate that neural machine translation models can be successfully applied to the task of error correction. 

While the experiments of this research are performed on an Arabic corpus, our methods in this thesis can be easily applied to any language.

   
\keywords{natural language error correction, recurrent neural networks, encoder-decoder models, attention mechanism}

\clearpage

%
%

\tableofcontents
\listoffigures
\addcontentsline{toc}{chapter}{List of Figures}
\listoftables
\addcontentsline{toc}{chapter}{List of Tables}


\newacronym{hmm}{HMM}{Hidden Markov Models}
\newacronym{crf}{CRF}{Conditional Random Field}
\newacronym{smt}{SMT}{Statistical Machine Translation}
\newacronym{nmt}{NMT}{Neural Machine Translation}
\newacronym{ann}{ANN}{Artificial Neural Networks}
\newacronym{mlp}{MLP}{Multilayer Perceptron}
\newacronym{rnn}{RNN}{Recurrent Neural Network}
\newacronym{sgd}{SGD}{Stochastic Gradient Descent}
\newacronym{brnn}{BRNN}{Bidirectional Recurrent Neural Network}
\newacronym{lstm}{LSTM}{Long Short-Term Memory}
\newacronym{gru}{GRU}{Gated Recurrent Unit}
\newacronym{bptt}{BPTT}{Back-Propagation Through Time}
\newacronym{qalb}{QALB}{Qatar Arabic Language Bank}

\glsaddall
\printindex
\printglossary[title=List of Abbreviations]
\clearpage
\renewcommand{\thepage}{\arabic{page}}
\chapter{Introduction}

\section{Motivations}

Automatic spelling and grammar correction is the task of automatically correcting errors in written text. If you type an incorrect word in a text environment in a phone or on text-editor software, it would be detected as an error, which means that the word is not the best choice for that context. It is then auto-corrected using another word, or a list of possible alternative words is suggested. Nowadays, error correction systems are inseparable components of any text-related application.

A few months ago, a tweet from Donald Trump went viral: ``Despite the constant negative press \textit{covfefe}``. Based on our prior knowledge from the context, we can guess that he meant to use the word "coverage" rather than "\textit{covfefe}". But how can this prior knowledge be represented in a computer system?

We can look at the problem of error correction in multiple ways. In traditional methods, correction models were generally based on the linguistic nature of errors. However, because of the notorious complexity and irregularity of human language, and the great variability in the types of error as well as their syntactic and semantic dependencies on the context, more models that performed better were needed. Example \ref{example1} demonstrates examples of error types that seem to need more complex methods than those that are language-based.

\begin{testexample}[Different types of errors]
\begin{itemize}[noitemsep]
\item We went to the store and bought \textit{new stove}. [new stove $\rightarrow$ a new stove]
\item Mary wishes that she \textit{does} not live in a dormitory. [\textit{does} $\rightarrow$ did] 
\item Anni said that he wants \textit{too} give her a gift. [ \textit{too} $\rightarrow$ to] 
\item He was so quiet that \textit{hardly he noticed} her.  [\textit{hardly he noticed} $\rightarrow$ he hardly noticed] 
\item I drank a double espresso. Now I'm \textit{entirely} awake. [entirely $\rightarrow$ wide] 
\end{itemize}
\label{example1}
\end{testexample}

Given a potentially erroneous input phrase, some approaches use classifiers to generate corrections by modeling their interactions with, for example, an n-gram \cite{ullmann1977binary} or a Conditional Random Fields model (CRF) \cite{mccallum2012conditional}. Statistical Machine Translation (SMT) systems have been used successfully in this context, in particular as a result of the increasing availability of manually annotated corpora. However, their major drawback is the difficulty of modeling corrections in different granularities, e.g., characters and words, which is necessary to reduce the rate of unknown words that are detrimental to their proper functioning. More recently, the use of neural networks has delivered significant gains for mapping tasks between pairs of sequences, including translation and spelling correction, due to to their ability to learn a better representation of the data and a better consideration of the context. 

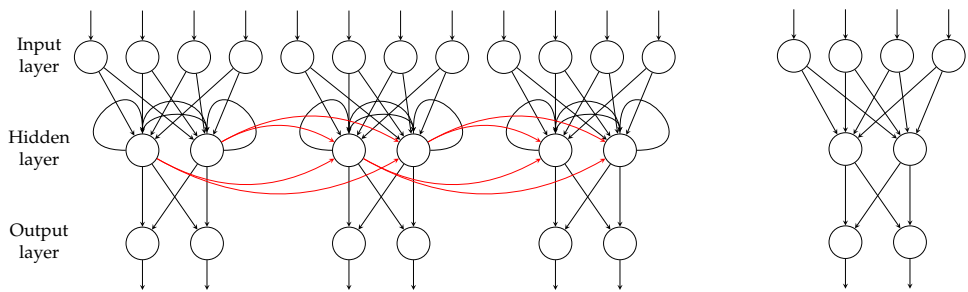
\begin{figure}[h]
    \centering
    \scalebox{0.95}{
        \begin{minipage}{0.65\textwidth}
        \centering
        \begin{center}
        \scalebox{.65}{
\begin{tikzpicture}[
   x={(0,-1cm)},
   y={(1.1cm,0)},
   >=stealth,
   ]
\foreach [count=\xx] \X in {0,4,8}{
\foreach \m/\l [count=\y] in {1,2,3,4}
  \node [every neuron/.try, neuron \m/.try] (input-\m-\xx) at (0,2.5-\y+\X) {};

\foreach \m [count=\y] in {1,2}
  \node [every neuron/.try, neuron \m/.try ] (hidden-\m-\xx) at (2,2-\y*1.25+\X) {};

\foreach \m [count=\y] in {1,2}
  \node [every neuron/.try, neuron \m/.try ] (output-\m-\xx) at (4,2-\y*1.25+\X) {};

\foreach \l [count=\i] in {1,2,3,n}
  \draw [<-] (input-\i-\xx) -- ++(-1,0);


\foreach \l [count=\i] in {1,n}
  \draw [->] (output-\i-\xx) -- ++(1,0);

\foreach \i in {1,...,4}
  \foreach \j in {1,...,2}
    \draw [->] (input-\i-\xx) -- (hidden-\j-\xx);

\foreach \i in {1,...,2}
  \foreach \j in {1,...,2}
    \draw [->] (hidden-\i-\xx) -- (output-\j-\xx);

\draw[->,shorten >=1pt] (hidden-1-\xx) to [out=0,in=90,loop,looseness=8.8] (hidden-1-\xx);
\draw[->,shorten >=1pt] (hidden-2-\xx) to [out=180,in=90,loop,looseness=8.8] (hidden-2-\xx);

\draw[->,shorten >=1pt] (hidden-1-\xx) to [out=90,in=90,loop,looseness=1.7] (hidden-2-\xx);
\draw[->,shorten >=1pt] (hidden-2-\xx) to [out=90,in=90,loop,looseness=1] (hidden-1-\xx);

}

\foreach \l/\txt [count=\x from 0] in {Input/input, Hidden/hidden, Output/output}
  \node [align=center] at (\txt-1-\xx-|0,-2.5) {\l \\ layer};

\foreach [evaluate=\i as \j using int(\i+1)] \i in {1,2}
{
  \draw [red,->] (hidden-1-\i) to[bend left] (hidden-1-\j);
  \draw [red,->] (hidden-1-\i) to[bend left] (hidden-2-\j);
  \draw [red,->] (hidden-2-\i) to[bend right] (hidden-1-\j);
  \draw [red,->] (hidden-2-\i) to[bend right] (hidden-2-\j);

}

\end{tikzpicture}
}
\end{center}
    \end{minipage}

    \begin{minipage}{.3\textwidth}
    \centering
    \begin{turn}{-90}
    \scalebox{.65}{
        \begin{tikzpicture}[x=1cm, y=1.1cm, >=stealth]

\foreach \m/\l [count=\y] in {1,2,3,4}
  \node [every neuron/.try, neuron \m/.try] (input-\m) at (0,2.5-\y) {};

\foreach \m [count=\y] in {1,2}
  \node [every neuron/.try, neuron \m/.try ] (hidden-\m) at (2,2-\y*1.25) {};

\foreach \m [count=\y] in {1,2}
  \node [every neuron/.try, neuron \m/.try ] (output-\m) at (4,2-\y*1.25) {};

\foreach \l [count=\i] in {1,2,3,n}
  \draw [<-] (input-\i) -- ++(-1,0)
    node [above, midway] {};

\foreach \l [count=\i] in {1,n}
  \node [above] at (hidden-\i.north) {};

\foreach \l [count=\i] in {1,n}
  \draw [->] (output-\i) -- ++(1,0)
    node [above, midway] {};

\foreach \i in {1,...,4}
  \foreach \j in {1,...,2}
    \draw [->] (input-\i) -- (hidden-\j);

\foreach \i in {1,...,2}
  \foreach \j in {1,...,2}
    \draw [->] (hidden-\i) -- (output-\j);

\end{tikzpicture}
}
\end{turn}
    \end{minipage}%
}
\caption{A recurrent network (in left) vs. a feed-forward network (in right). }
\label{MLP_vs_rnn}
\end{figure}

In a simple feed-forward neural network, such as Multi-Layer Perceptron (MLP), we assume that all inputs and outputs of the network are independent on each other, which is not the best method for many tasks, particularly for language modeling. By sharing parameters across different parts of a model, we can turn an MLP into a Recurrent Neural Network (RNN). RNNs are a set of neural networks for processing sequential data and modeling long-distance dependencies which is a common phenomenon in human language. The simplest form of a recurrent neural network is an MLP with the previous set of hidden unit activations feeding back into the network together with the inputs, so that the activations can follow a loop. Therefore, unlike feed-forward networks --- which are amnesiacs regarding their recent past --- the MLP enables the network to do temporal processing and learn sequences, which are important features for language modeling. Figure \ref{MLP_vs_rnn} illustrates a feed-forward network and a recurrent network.

In theory, RNNs are capable of handling long-term dependencies, but this is not the case in practice. Different architectures have been proposed to tackle this problem, from which we have used Long Short Term Memory (LSTM) architecture. LSTMs help maintain a constant error that can be back-propagated through time and layers, so that they allow recurrent networks to continue learning gradually over time. Henceforth, our implementation of an RNN model is based on the LSTM architecture.

\begin{figure}[h]
\centering
\scalebox{.8}{

\begin{tikzpicture}[
  hid/.style 2 args={
    rectangle split,
    rectangle split horizontal,
    draw=#2,
    rectangle split parts=#1,
    fill=#2!20,
    outer sep=1mm}]
  \foreach \i [count=\step from 1] in {thr,unpresidented,attaker,{{$<$eos$>$}}}
    \node (i\step) at (2*\step, -2) {\i};
  \foreach \t [count=\step from 4] in {the,unprecedented,attacker,{{$<$eos$>$}}} {
    \node[align=center] (o\step) at (2*\step, +2.75) {\t};
  }
  \foreach \step in {1,...,3} {
    \node[hid={3}{cyan}] (h\step) at (2*\step, 0) {};
    \node[hid={3}{cyan}] (e\step) at (2*\step, -1) {};    
    \draw[->] (i\step.north) -> (e\step.south);
    \draw[->] (e\step.north) -> (h\step.south);
  }
  \foreach \step in {4,...,7} {
    \node[hid={3}{green}] (s\step) at (2*\step, 1.25) {};
    \node[hid={3}{black}] (h\step) at (2*\step, 0) {};
    \node[hid={3}{black}] (e\step) at (2*\step, -1) {};    
    \draw[->] (e\step.north) -> (h\step.south);
    \draw[->] (h\step.north) -> (s\step.south);
    \draw[->] (s\step.north) -> (o\step.south);
  }  
  \draw[->] (i4.north) -> (e4.south);
  \foreach \step in {1,...,6} {
    \pgfmathtruncatemacro{\next}{add(\step,1)}
    \draw[->] (h\step.east) -> (h\next.west);
  }
  \foreach \step in {4,...,6} {
    \pgfmathtruncatemacro{\next}{add(\step,1)}
    \path (o\step.north) edge[->,out=45,in=225] (e\next.south);
  }
\end{tikzpicture}
}
\caption{Encoder-Decoder model. Encoder units are specified in cyan and decoder units in gray. Green rectangles refer to the Softmax function, used for a probabilistic interpretation for the output.}
\label{encoder_decoder}
\end{figure}
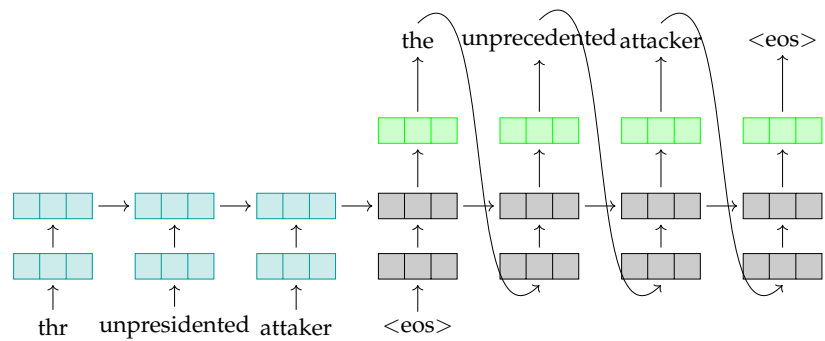

In this project, we are particularly interested in Encoder-Decoder models. The basic idea of the model is relatively simple. We use two RNN models: the first encodes the information of the potentially erroneous input text, as a vector of real-value numbers, while the second model decodes this information into the target sentence which is the corrected prediction for the input text. Figure \ref{encoder_decoder} demonstrates an encoder-decoder model for error correction.

The fact that the encoder-decoder attempts to store whole sentences of any arbitrary length in a hidden vector of fixed size, leads to a substantial size for the network which hinders the training process for large data sets. On the other hand, even if the network were large, it would not be efficient in terms of memory and time when processing shorter phrases. In addition, the long-distance dependencies would be ignored by time, even if LSTM were supposed to correct it. To remedy this problem, an attention mechanism is used to allow the decoder to \emph{attend} different parts of the source sentence at each step of the output generation.

The ability to correct errors accurately will improve the reliability  of the underlying applications and thus numerous commercial and academic implications. It facilitates the construction of software to help foreign language learning, as it reduces noise in the entry of NLP tools, thus improving their performance, especially on unedited texts that can be found on the Web.

\section{Contributions}

Inspired by the prevailing successes of neural networks, we decided to explore in detail certain models for the task of spelling and grammar correction. This thesis focuses on how to create such systems using sequence-to-sequence models together with attention mechanism. The first of two contributions is the architecture and training of such a model at character-level. The ultimate aim is to explore whether NMT methods can deliver competitive results. 

Although our research does not aim at a specific language, since models are tested on an Arabic corpus, this thesis also explores whether this project does contribute better error correcting than previous works for Arabic. To summarize, the following tasks have been performed in chronological order:

\begin{itemize}[noitemsep]
\item Collection and preprocessing data.
\item Implementation of our target models using DyNet. Although similar models were already implemented in other toolkits, e.g., Theano \footnote{\url{http://www.deeplearning.net/software/theano/}}, TensorFlow \footnote{\url{https://www.tensorflow.org/}}, we had to code the models entirely in DyNet \footnote{\url{https://github.com/clab/dynet}}. 
\item Experiments and comparison of our results with the existing results
\item Documentation of the research project
\end{itemize}

\section{Thesis Outline}

This introductory chapter is followed by a review in chapter \ref{Related_work} of previous research on the same theme. Chapter \ref{Background} describes the background methods needed for this research. In chapter \ref{Experimental_Setup}, the experimental setup of the work is described. In this section the corpus is preprocessed and data sets are created. Chapter \ref{Method} describes further details about the models that are used for error correction. In chapter \ref{Evaluation}, the results of our correction system are presented. This work is concluded in chapter \ref{Conclusion}. We will also propose several new ideas for future works in this domain. The final sections of this thesis are devoted to the appendices in \ref{appendix} and the bibliography in \ref{reference}.

\section{Target Audience}

This thesis endeavors to provide clear and coherent information for two audiences: developers and computational linguists. Since this work does not address the linguistic aspects of spelling and grammatical error correction, it may not be useful for linguistic researchers.

\chapter{Related work}
\label{Related_work}

This chapter reviews previous research on automatic spelling and grammar correction. Section \ref{error_detection} presents a general survey on the methods often used for error detection tasks. Following this section, section \ref{error_correction} explains the most common error correction techniques. Since our principal task is predominantly related to error correction using neural networks, the emphasis in chapter \ref{Background} is on introducing the basic notions of neural networks, particularly Recurrent Neural Networks.

\section{Error detection techniques}
\label{error_detection}

In most systems, before any correction is carried out, a detection process is conducted on the input sentence to extract the potentially incorrect words. Two main techniques \cite{kukich1992techniques} are used to detect non-word spelling errors that are explained in the following sections. A non-word error refers to a potentially incorrect word that does not exist in a given dictionary.

\subsection{Dictionary lookup}

Dictionary lookup is one of the basic techniques employed to compare input strings with the entries of a language resource, e.g., lexicon or corpus. Such a language resource must contain all inflected forms of the words and it should be updated regularly. If a given word does not exist in the language resource, it will be marked as a potentially incorrect word. Reducing the size of resources and ameliorating the search performance, through morphological analysis and pattern-matching algorithms (e.g., hashing, search trees), presents a challenge in this method.

\subsection{n-gram analysis}
\label{n_gram}

Statistical models were designed to assign a probability to a sequence of symbols, e.g. characters and words. \textit{n-gram} is one of the popular models that represents an attempt to count a structure by counting its substructure. Given a sentence $C=c_1,c_2,...,c_l=c_1^l$, we can define the probability over a character $c_i$ as follows:

\begin{equation}
P(C=c_1,c_2,...,c_l=c_1^l) = \prod_{i=1}^{l} P(c_i|c_1,c_2,...,c_{i-1})
\end{equation}

Assuming different $n$-values in the $n$-gram model, we can create different probabilistic models. In a unigram model ($n=1$), the probability of each character is calculated independently from previous characters, i.e., $P(c_i|c_1,c_2,...,c_{i-1}) \approx P(c_i)$. Although the unigram model ignores the context by taking only the current character $c_i$ into account, the models where $n>1$ can represent a more accurate probability of the characters, since $P(c_i|c_1,c_2,...,c_{i-1}) \approx P(c_i|c_{i-n},...,c_{i-1})$. A simple way to estimate probabilities for each character is known as maximum likelihood estimation (MLE). Equation \ref{equationMLE} calculates the n-gram probability by dividing the number of times a particular string is observed by the frequency of the context.

\begin{equation}
P(c_i| c_{i-n+1}^{i-1})=\frac{count(c_{i-n+1}^i)}{count(c_{i-n+1}^{i-1})}
\label{equationMLE}
\end{equation}

For the task of error detection, n-gram analysis estimates the likelihood that a given input will be correctly spelled. To accomplish this, a pre-calculated n-gram statistic from the language resource is provided for comparative tasks. The choice of $n$ depends on the size of the training corpus.

\section{Error correction techniques}
\label{error_correction}

The task of machine correction is defined as correcting a $N$-character source sentence $S = s_1,..., s_N = s_1^N$ into a $M-character$ target sentence $T = t_1,...,t_M = t_1^M$. Thus, any type of correction system can be defined as a function $MC$:

\begin{equation}
\widehat{T}=MC(S)
\label{equation:1}
\end{equation}

which returns a correction hypothesis $\widehat{T}$ given an input word $S$. In the case of word-level models, inputs can be considered as the source words and the output are the predicted words based on the same function. 

\subsection{Minimum edit distance}

Minimum edit distance is one of the most studied techniques for error correction. It is based on counting editing operations, which are defined in most systems as insertion, deletion, substitution and transposition, in order to transform an incorrect input into the most probable word, i.e., the one with least edit distance. Hamming \cite{hamming1950error}, Jaro–Winkler \cite{winkler1990string}, Wagner–Fischer \cite{wagner1974string}, Damerau-Levenshtein \cite{Damerau:1964:TCD:363958.363994} and Levenshtein \cite{levenshtein1966binary} are among the most famous edit distance algorithms. We will Levenshtein distance later in one of the evaluation metrics in section \ref{m2_method}.

\subsection{Similarity key technique}

Similarity key technique is another technique for error correction. It is based on classifying characters into the groups with similar keys. When a word is detected to be potentially incorrect, its characters are mapped into the pre-defined keys so that all other words in the language resource that have the same key are suggested as correction candidates. The candidates are then ranked according to the minimum edit distance.

Various algorithms propose different approaches to design key groups for characters of a language. In general, the similarity is measured based on the position and the order of the characters in the words. Anagrams and the phonology of language are suitable factors to construct corresponding keys for each character. Soundex \cite{davidson1962retrieval}, Metaphone \cite{clm/philips90} and SPEEDCOP \cite{pollock1983collection} and Caverphone \cite{phua2006personal} are among the most common methods using similarity key technique.

\subsection{Rule-based techniques}

By analyzing the most common spelling errors, some researchers have attempted to create a knowledge base of errors for the task of correction \cite{yannakoudakis1983rules,means1988cn,Russell:2003:AIM:773294} using rule-based models that encode grammatical knowledge. These rules are generally based on the morphological characteristics of the language.

\subsection{Probabilistic techniques}

Approaches based on stochastic models that learn a suitable representation from data have been also used for error correction tasks. Statistical machine correction systems, like the similar systems in machine translation \cite{Manning:1999:FSN:311445}. They perform their task by creating a probabilistic model on sequences of symbols, e.g., characters and words, also known as \emph{language model}, in such a way that the desired output has the highest probability giving specific parameters of the model $\theta$. Therefore, we have:

\begin{equation}
\widehat{T} = \underset{T}{argmax}P(T|S;\theta) 
\label{eq3.2}
\end{equation}

which is called the Fundamental Equation of Machine Translation \cite{brown1993mathematics}. 

Various approaches are used to model the probability. n-gram-based language models, as mentioned in section \ref{n_gram}, often also function as a probabilistic technique for error correction. Another common method is the log-linear language model \cite{rosenfeld1996maximum} which calculates the probability by creating a feature vector that describes the context using different features and then calculating a score vector that corresponds to the likelihood of each symbol.

More highly complex probabilistic network representations have been also introduced \cite{Manning:1999:FSN:311445,Koller:2009:PGM:1795555}. Hidden Markov Models (HMM), for example, have demonstrated a strong ability to model human language \cite{lottaz2003modeling}, ut since HMMs assume conditional independence of the previous words except the last one, they are not a practicable choice for modeling the long-distance dependencies. On the other hand, the recurrent neural networks (RNN) have been demonstrated to be more capable of modeling such dependencies \cite{DBLP:journals/corr/KarpathyJL15}. 

Neural networks are also based on the probability distribution of language and they have shown recent success in different tasks related to natural language processing. This study focuses on attention-based encoder decoder neural networks for error correction. Such systems consist of two components: an encoder that computes a representation for an input sentence, and a decoder that generates one target word at a time and decomposes the conditional probability as

\begin{equation}
log(p(T|S)) = \sum_{j=1}^{m}log p(t_j;t_{<j},S)
\end{equation}

where $t_j$ is the current word of sentence $T$, $t_{<j}$ refers to the previous words and $S$ is the source sentence. 

These neural machine translation (NMT) systems have been used in recent related works with different architectures of the decoder and different representations in the encoder. Kalchbrenner and Blunsom \cite{DBLP:journals/corr/KalchbrennerB13} used an RNN for the decoder and a convolutional neural network for encoding the source sentence.  Sutskever et al. \cite{sutskever2014sequence} and Luong et al. \cite{luong2015effective}, on the other hand, have used multiple layers of an RNN with a Long Short-Term Memory (LSTM) hidden unit for both encoder and decoder. Cho et al. \cite{cho2014learning}, Bahdanau et al. \cite{bahdanau2014neural} and Jean et al. \cite{jean2014using} have all used another architecture for hidden units, the Gated Recurrent Unit (GRU), for both encoding and decoding components. 

The work of Luong et al. \cite{luong2015effective} is the most relevant to the present study. While our aim is to apply the same principals to error correction task in character-level models. We can also say that this work is the continuation of the generalized character-level spelling error correction presented in \cite{Farra2014GeneralizedCS} which is a model that maps input characters into output characters using supervised learning. Although the model is applied to correct errors in Arabic, it does not consider morphological or linguistic features. On the other hand, the model is context-sensitive, and it explores beyond the context of the given word.

\chapter{Background}
\label{Background}

This chapter is divided into two sections and provides an overview of the background needed for constructing our error correction system: section \ref{section_nn} explains the most important characteristics of neural networks and special methods that can be used for modeling long-distance dependencies. Section \ref{evaluation_metrics} then addresses the most common evaluation metrics for error correction systems. The evaluation results of our designed models in chapter \ref{Method} are analyzed in chapter \ref{Evaluation}.

\section{Neural Networks}
\label{section_nn}

Artificial neural networks (ANN) were originally developed as mathematical models in-spired by the biological neural networks. A biological neural network is an interconnected network of neurons transmitting elaborate patterns of electrical signals. Dendrites receive input signals and, based on those inputs, trigger an output signal via an axon. An ANN follows the same principle using mathematical computations to process inputs.

In 1943, McCulloch and Pitts modeled mathematically the biological neural networks \cite{mcculloch1990logical}. This network could solve binary problems but without learning. Later, in 1957, Rosenblatt presented the first ANN that was able to vary its own weights, in other words, the network could learn and ameliorate itself (Figure \ref{fig_perceptron}).

\def\layersep{1.5cm}

\begin{figure}[h]
\centering
\scalebox{0.9}{
\begin{minipage}{.45\linewidth}

\scalebox{0.9}{
\centering
\begin{tikzpicture}
        \node[functions] (center) {};
        \node[above of=center,font=\scriptsize,text width=4em] {Activation function};
        \draw[thick] (0.5em,0.5em) -- (0,0.5em) -- (0,-0.5em) -- (-0.5em,-0.5em);
        \draw (0em,0.75em) -- (0em,-0.75em);
        \draw (0.75em,0em) -- (-0.75em,0em);
        \node[right of=center] (right) {};
            \path[draw,->] (center) -- (right);
        \node[functions,left=3em of center] (left) {$\sum$};
            \path[draw,->] (left) -- (center);
        \node[weights,left=3em of left] (2) {$w_2$} -- (2) node[input,left of=2] (l2) {$x_2$};
            \path[draw,->] (l2) -- (2);
            \path[draw,->] (2) -- (left);
        \node[below of=2] (dots) {$\vdots$} -- (dots) node[left of=dots] (ldots) {$\vdots$};
        \node[weights,below of=dots] (n) {$w_n$} -- (n) node[input,left of=n] (ln) {$x_n$};
            \path[draw,->] (ln) -- (n);
            \path[draw,->] (n) -- (left);
        \node[weights,above of=2] (1) {$w_1$} -- (1) node[input,left of=1] (l1) {$x_1$};
            \path[draw,->] (l1) -- (1);
            \path[draw,->] (1) -- (left);
        \node[weights,above of=1] (0) {$w_0$} -- (0) node[input,left of=0] (l0) {$1$};
            \path[draw,->] (l0) -- (0);
            \path[draw,->] (0) -- (left);
        \node[above of=l0,font=\scriptsize] {inputs};
        \node[above of=0,font=\scriptsize] {weights};
        
    \end{tikzpicture}
    }
\end{minipage}
\hspace{.05\linewidth}
\centering
\begin{minipage}{.45\linewidth}
\scalebox{0.9}{
  \begin{tikzpicture}[shorten >=1pt,->,draw=black!50, node distance=\layersep]
    \tikzstyle{every pin edge}=[<-,shorten <=1pt]
    \tikzstyle{neuron}=[circle,fill=black!25,minimum size=17pt,inner sep=0pt]
    \tikzstyle{input neuron}=[neuron, fill=green!50];
    \tikzstyle{output neuron}=[neuron, fill=red!50];
    \tikzstyle{hidden neuron}=[neuron, fill=cyan!50];
    \tikzstyle{annot} = [text width=4em, text centered]

    \foreach \name / \y in {1,...,4}
        \node[input neuron] (I-\name) at (0,-\y) {$x_{\y}$};

    \foreach \name / \y in {1,...,5}
        \path[yshift=0.5cm]
            node[hidden neuron] (H-\name) at (\layersep,-\y cm) {};

    \node[output neuron,pin={[pin edge={->}]right:Output}, right of=H-3] (O) {};

    \foreach \source in {1,...,4}
        \foreach \dest in {1,...,5}
            \path (I-\source) edge (H-\dest);

    \foreach \source in {1,...,5}
        \path (H-\source) edge (O);

    \node[annot,above of=H-1, node distance=1cm] (hl) {Hidden layer};
    \node[annot,left of=hl] {Input layer};
    \node[annot,right of=hl] {Output layer};
\end{tikzpicture}
}
\end{minipage}
}
	\caption{A simple Perceptron (in left) vs. a Multi-Layer Perceptron (in right)}
	\label{fig_perceptron}
\end{figure}
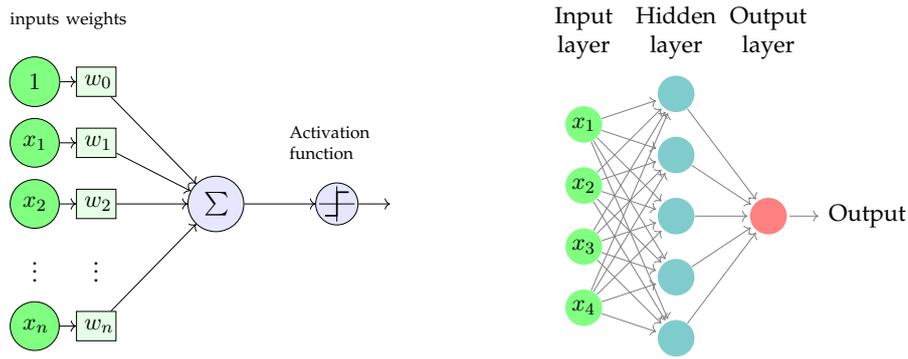

The perceptron computes a single output from multiple real-valued inputs by forming a linear combination according to its input weights and then possibly putting the output through some nonlinear activation function. Technically speaking, let $x_1,x_2,...,x_n$  $\epsilon$ be $n$ scalar inputs of the network. Output $y$, as function of a sum of weighted inputs is computed by:

\begin{equation}
z = \sum_{i = 1}^{n}w_ix_i+b = w^Tx + b
\label{affine_tran}
\end{equation}

where w=$(w_1, w_2,...,w_n)$  $\epsilon$ is corresponding weights, $b$ $\epsilon$ is bias and $\varphi$ is an activation function. The weight is often referred to as \textit{pre-activation} and the bias is replaced by an equivalent input term $x_0=1$ weighted by $w_0=b$. Note that this is simply the dot product of the weight vector w and the input vector x. The result of this affine transformation $z$ is then passed through a step function that determines the binary output of the Perceptron. 

\begin{equation}
y = 
\left\{\begin{matrix}
1  & if z \geq  0\\ 
0 & otherwise
\end{matrix}\right.
\end{equation}

Given a training pair (x, y), the parameters of the network, i.e., weights and bias, are learning using the following rule:

\begin{equation}
w_i \leftarrow w_i - \eta.(\widehat{y}-y).x_i
\end{equation}

where $\widehat{y}$ is the output of the Perceptron, $y$ is the desired output and $\eta$ is the learning rate to adjust the updating magnitude. Note that the $\leftarrow$ refers to the assignment of the new value of $w_i$.

\subsection{Multilayer Perceptron}
Although conceptually remarkable, Perceptron is significantly limited in modeling nonlinearly separable problems. A famous example is that of the XOR logic operation which cannot be solved by a linear transformation. An evolution of Perceptrons introduces an intermediate between the input and the output, which is known as the hidden layer. Hidden layers allow the pre-activation to be followed by a nonlinearity, which produces a nonlinear transformation that can map the input into a linearly separable space, such as the case of XOR. Figure \ref{fig_perceptron} demonstrates an MLP.

\begin{figure}[h]
    \centering
    \begin{tikzpicture}[domain=-2.9:2.9]
        \draw[very thin,color=gray] (-2.9, -1.1) grid (2.9, 2.9);
        \foreach \x in {-2,-1,0,1,2}
            \node[anchor=north] at (\x,-1.1) {\x};
        \foreach \y in {-1,0,1,2}
            \node[anchor=east] at (-2.9,\y) {\y};
        \draw[->] (-2.9, 0) -- (3.1, 0) node[right] {$z$};
        \draw[->] (0, -1.1) -- (0, 3.1) node[above] {$f(z)$};

        \draw[thick, color=blue]
            plot (\x, { 1/(1+exp(-1 * \x)) })
            node[right,yshift=-0.5em] {$logistic(z)$};
        \draw[thick, color=black]
            plot (\x, { ( 1 - exp(-2*\x) ) / ( 1 + exp(-2*\x) ) })
            node[right,yshift=0.5em] {$tanh(z)$};
        \draw[thick, color=red]
            plot (\x, { max(0, \x) })
            node[xshift=-19.3em,yshift=-8em] (a) {$ReLU(z)$};
    \end{tikzpicture}
    \caption{Most common activation functions: sigmoid, tanh, ReLU.}
        \label{fig:activations}
\end{figure}
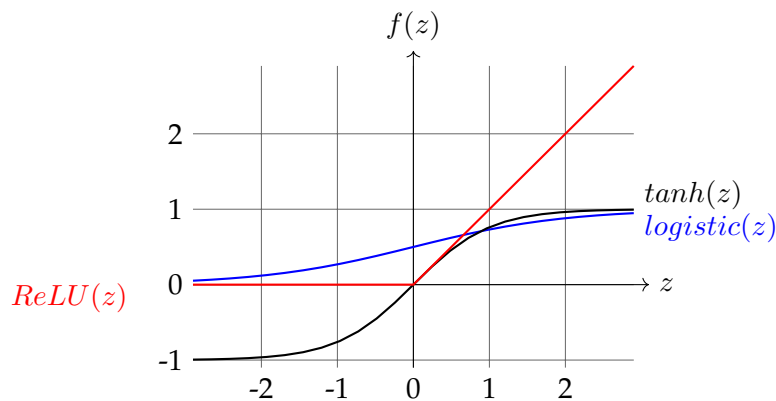

An MLP is a network of Perceptrons grouped in interconnected layers. Each connection from the $i$-th neuron of layer $l$-1 to the $j$-th neuron of layer $l$ is associated to a weight $w_{ij}^{(l)}$, and all the weights are stored in a matrix $W^{(l)}$. It is a feed-forward neural network with one or more hidden layers between the input and output layer. Feed-forward means that data flows in one direction from input to output layer. Similar to the Perceptron, each layer computes an affine transformation as follows:

\begin{equation}
z^{(l)} = W^{(l)}.a^{(l-1)} 
\end{equation}

Generally, each layer of an ANN computes some activation based on its input and a nonlinear activation function. The choice of which activation function to use in each layer can have a significant impact on the performance of the model. An activation function is a non-linear function that maps any real-valued number to a reduced range. Hyperbolic tangent (tanh), logistic function and rectified linear unit (ReLU) function are among the most common activation function. They do the aforementioned mapping in [-1, 1], [0, 1] and [0, $x$] respectively. These functions are illustrated in figure \ref{fig:activations}.

\begin{gather}
tanh(x) = \frac{e^{2x}-1}{e^{2x}+1}\\
\sigma (x) = \frac{1}{1+e^{-x}}\\
ReLU(x) = max(0, x)
\end{gather}

Both the number of units in the output layer and the choice of output activation function depend on the task. Binary classification can be performed with a one-neuron output layer. An output can be interpreted as the probability $Pr(True|x)$ of assigning a predefined class to an input $x$. In the case that a logistic sigmoid is used as the activation function, the model is called \textit{logistic regression}, or a \textit{logit} model. In a more general way, we can solve the same problematic for an N-class classification ($N > 2$) task, by using N output neurons normalizing by a softmax function \cite{costa1996probabilistic}. The softmax is a squashing function that maps its input to a categorical distribution. It is defined as follows:

\begin{equation}
\underset{i}{softmax(z)} = \frac{exp(z_i)}{\sum_{n=0}^{N}exp(z_n)}
\end{equation}

where $z$ refers to the affine transformation mentioned in equation \ref{affine_tran}. 
\subsubsection{Back-propagation}

Since the solution of the network exists in the values of the parameters (weights and bias)  $\theta$ between neurons, an optimization process is needed to acquire parameters that fit the training corpus. One of the most common algorithms is termed back-propagation. The basic idea is to modify each weight of the network by a factor proportional to the error, i.e., the difference between the desired output and the output of the network. Different functions have been introduced as objective function to calculate the error from which we use the Cross-Entropy cost as follows:

\begin{equation}
H(\widehat{y}-y) = \sum_{i} y_ilog\frac{1}{\widehat{y}_i} = - \sum_{i} y_ilog\widehat{y}_i
\label{equation_cross_entropy}
\end{equation}

where $\widehat{y}_i = MLP(x_i;\theta)$ refers to the output of the network and $y$ is the desired output. The summation computes the overall loss of the network on a single pair. An objective function is either a loss function or its negative form. Here, we assume that equation \ref{equation_cross_entropy} is our objective function. In fact, we calculate the error in order to reduce it for each pair by searching for the best parameters $\theta$:

\begin{equation}
\theta^* = \underset{\theta}{argmin}H(\widehat{y}-y)
\end{equation}

Since MLPs are differentiable operators, except for the case where ReLU function is used, they can be trained to minimize any differentiable cost function using \textit{gradient descent}. The gradient descent aims to determine the derivative of the objective function with respect to each of the network weights, then adjust the weights in the direction of the negative slope. Having said that since the ReLU function is not differentiable, Leaky ReLU is generally used for back-propagation where:

\begin{equation}
LeakyReLU(x) = \left\{\begin{matrix}
x & \\ if x > 0
0.01x & otherwise
\end{matrix}\right.
\end{equation}

\paragraph{Stochastic Gradient Descent} 

Stochastic Gradient Descent (SGD) is an iterative process where we pick an observation uniformly at random, say $i$ and attempt to improve the likelihood with respect to the observation. Each iteration, also called epoch, $\theta$ is updated as follows:

\begin{equation}
\theta \leftarrow \theta - \eta \triangledown _{x_i}H(\widehat{y}-y)
\end{equation}

where $\eta$ is the learning rate that determines the size of the steps needed to reach a local minimum. In other words, the gradient follows the slope direction of the surface created by the objective function until it reaches a minimum.

While SGD performs one update at a time, two variants of SGD perform this update differently. In the Vanilla Gradient Descent method, the gradient of the loss function is computed for the entire training set. Thus, it can be extremely slow and intractable for large training sets. On the other hand, the Mini-batch Gradient Descent --- instead of calculating the entire training set --- divides it into small subsets and computes the gradient with respect to a subset of the training set. In other words, this model uses the efficiency of the Vanilla Gradient Descent in finding the optimal convergence and the performance of the stochastic variant.

In appendix \ref{adam} we will introduce Adaptive Moment Estimation (Adam) optimization method as well. 

\subsection{Recurrent Neural Networks}

Recurrent neural network (RNN) is another type of neural network that is particularly used for language processing tasks. The main characteristic of an RNN, in addition to those of any ANN (e.g., units and parameters), is the cyclic connections that each unit can have. This provides a memory-like functionality for the RNN which could be an appropriate model for language processing. Unlike a simple MLP that maps inputs into the outputs directly, an RNN can make use of, theoretically, all the previous inputs for each output, thus providing an internal state other than the network's parameters.

In order to create an RNN, we first need to compress a sequence of input symbols $X=(x_1,x_2,...,x_n)$ into a fixed-dimensional vector by using recursion. Using this vector enables us to deal with variable-length inputs and outputs. Then, assume at step $t$ that we have a vector $h_{t-1}$ which is the history of all previous symbols before the current one. The RNN will compute the new vector, or its internal state, $h_t$ which compresses all the previous symbols $\left( x_1, x_2, \dots, x_{t-1} \right)$ as well as the new symbol $x_t$ by

\begin{gather}
h_t=tanh(Wx_t+Uh_{t-1}+b)\\
\widehat{y}_t=softmax(Vh_t)
\label{rnn_equation_1}
\end{gather}

where $W$ is the input weight matrix, $U$ is the recurrent weight matrix, $V$ is the weight of the hidden layer and $b$ is the bias vector. Although this is not the only way to model an RNN \cite{DBLP:journals/corr/PascanuGCB13}, we have used this formulation in our implementation (which is known as Elman network \cite{lenneberg1967biological}).

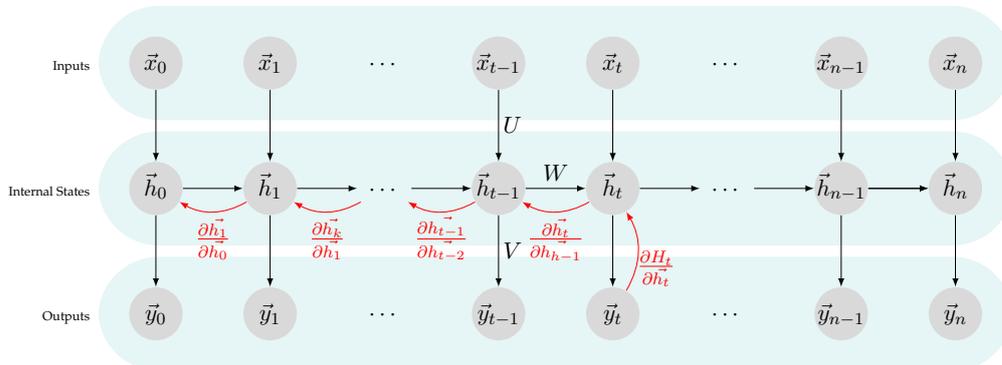
\begin{figure}[h]
\tikzstyle{background}=[rectangle,fill=cyan!10,inner sep=0.5cm, rounded corners=10mm]
\scalebox{0.8}{
\begin{tikzpicture}[>=latex,text height=1.5ex,text depth=0.25ex,
el/.style = {inner sep=2pt, align=left, sloped},
every label/.append style = {font=\tiny}]
	\tikzstyle{neuron}=[circle,fill=gray!25,minimum size=25pt,inner sep=0pt]
	\tikzstyle{cdots_neuron}=[circle,fill=cyan!10,minimum size=25pt,inner sep=0pt]
  

  \matrix[row sep=0.6cm,column sep=0.5cm,ampersand replacement=\&] {
    	\&
    	\node (x_0) [neuron]{$\vec{x}_0$}; \&
        \&
        \node (x_1) [neuron]{$\vec{x}_1$}; \&
        \&
		\node (cdots_x)   [cdots_neuron] {$\cdots$};	\&
        \&
        \node (x_t-1) [neuron]{$\vec{x}_{t-1}$}; \&
        \&
        \node (x_t)   [neuron]{$\vec{x}_t$};     \&
		\&
		\node (cdots_x)   [cdots_neuron] {$\cdots$};	\&
        \&
        \node (x_n-1)   [neuron] {$\vec{x}_{n-1}$};	\&
        \&         	
        \node (x_n) [neuron]{$\vec{x}_{n}$}; \&

        \\
		\\
		\&
        \node (s_0) [neuron] {$\vec{h}_0$}; \&
        \&
        \node (s_1) [neuron] {$\vec{h}_1$}; \&
        \&
		\node (cdots_s2)   [cdots_neuron] {$\cdots$};	\&
		\&
        \node (s_t-1) [neuron] {$\vec{h}_{t-1}$}; \&
		\&
        \node (s_t)   [neuron] {$\vec{h}_t$};     \&
        
		\&
		\node (cdots_s)   [cdots_neuron] {$\cdots$};
													\&
        	\&
        	\node (s_n-1)   [neuron] {$\vec{h}_{n-1}$};
													\&
        	\&        		
        \node (s_n) [neuron] {$\vec{h}_{n}$}; \&
		\\
		\\

        \&
        \node (o_0) [neuron] {$\vec{y}_0$}; \&
        \&
        \node (o_1) [neuron] {$\vec{y}_1$}; \&
        \&
		\node (cdots_x)   [cdots_neuron] {$\cdots$};	\&
        \&
        \node (o_t-1) [neuron] {$\vec{y}_{t-1}$}; \&
        \&
        \node (o_t)   [neuron] {$\vec{y}_t$};     \&
        	\&
		\node (cdots_o)   [cdots_neuron] {$\cdots$};	\&
        \&
        \node (o_n-1)   [neuron] {$\vec{y}_{n-1}$};		\&
        \&         	
        \node (o_n) [neuron] {$\vec{y}_{n}$}; \&
        \\
    };
    
    \path[->]
    	(x_0) edge (s_0)
  		 (x_1) edge (s_1)
    	(x_t-1) edge node[el,right,rotate=90]  {$U$}  (s_t-1)
        (x_t) edge (s_t)
        (x_n-1) edge (s_n-1)
        (x_n) edge (s_n)

        (s_t-1) edge node[el,above]  {$W$} (s_t)
        (s_t) edge (cdots_s)  
        (s_n-1) edge (s_n)
        (s_0) edge (s_1)
        (s_n-1) edge (s_n)

        (s_0) edge (o_0)
        (s_1) edge (o_1)
        (s_1) edge (cdots_s2)
        (cdots_s2) edge (s_t-1)
        (cdots_s) edge (s_n-1)
    	(s_t-1) edge node[el,right,rotate=90]  {$V$}  (o_t-1)
        (s_t) edge (o_t)
        (s_n-1) edge (o_n-1)
        (s_n) edge (o_n)
        
		(s_t)  edge[bend left, red] node[el,below,yshift=-.2cm] {$\frac{\partial \vec{h_t}}{\partial \vec{h_{h-1}}}$} (s_t-1)
       (o_t)  edge[bend right, red] node[el,below,yshift=-.3cm,rotate=-90] {$\frac{\partial H_t}{\partial \vec{h_{t}}}$} (s_t) 
       (s_t-1)  edge[bend left, red] node[el,below,yshift=-.2cm] {$\frac{\partial \vec{h_{t-1}}}{\partial \vec{h_{t-2}}}$} (cdots_s2) 
       (cdots_s2)  edge[bend left, red] node[el,below,yshift=-.2cm] {$\frac{\partial \vec{h_k}}{\partial \vec{h_{1}}}$} (s_1) 
       (s_1)  edge[bend left, red] node[el,below,yshift=-.2cm] {$\frac{\partial \vec{h_1}}{\partial \vec{h_{0}}}$} (s_0) 
        
        ;
    
    \begin{pgfonlayer}{background}
        \node [background,
                    fit=(x_0) (x_n),
                    label=left:Inputs] {};
        \node [background,
                    fit=(s_0) (s_n),
                    label=left:Internal States] {};
        \node [background,
                    fit=(o_0) (o_n),
                    label=left:Outputs] {};
    \end{pgfonlayer}
\end{tikzpicture}
}
  \caption{A Recurrent Neural Network with one hidden layer in different time steps. The gradient calculation for time step $t$ is shown in red.}
  \label{simple_rnn}
\end{figure}

Training an RNN is similar to training a traditional Neural Network. We also use the back-propagation algorithm, but with some modifications. Since the parameters are shared by all time steps in the network, the gradient at each output depends not only on the calculations of the current time step, but also the previous time steps. For instance, in order to calculate the gradient at $t=4$, we need to back-propagate three steps and sum up the gradients. This method is called \textit{Back-propagation Through Time} (BPTT). 

\subsubsection{Backpropagation Through Time}

In the previous section, we explained the basic mechanism of an RNN. In calculating the back-propagation, we mentioned that the gradients of the error should be calculated with respect to the parameters of the network, i.e., $U$, $V$ and $W$, for the current and previous time steps. To do so, we take use of the BPTT algorithm to do so.  

Let us assume that we want to calculate the gradients for time step $t$. To calculate the gradients, using the chain rule of differentiation, we have:

\begin{equation}
\frac{\partial H_t}{\partial V}= \frac{\partial H_t}{\partial \widehat{y_t}} \frac{\partial \widehat{y_t}}{\partial V} = \frac{\partial H_t}{\partial \widehat{y_t}} \frac{\partial \widehat{y_t}}{\partial z_t}\frac{\partial z_t}{\partial V} = (\widehat{y_t}-y_t) \otimes h_t
\end{equation}

where $z_t=Vh_t$ and $\otimes$ refers to the outer product of two vectors. As we see, $ \frac{\partial H_t}{\partial V}$ depends only on the values at the current time step $y_t$, $\widehat{y_t}$ and $h_t$. Although the gradient with respect to $W$ and $U$ is a little different:

\begin{equation}
\frac{\partial H_t}{\partial W} = \frac{\partial H_t}{\partial \widehat{y_t}}\frac{\partial \widehat{y_t}}{\partial h_t}\frac{\partial h_t}{\partial W}
\end{equation}

where $h_t$ is the activation function \ref{rnn_equation_1} which depends on the previous internal state $h_t-1$ which depends on the previous states $h_{t-2}, h_{t-3},...,h_0$ and $W$ (shown in figure \ref{simple_rnn} in red). So we need to apply the chain rule on all the previous states to obtain the gradient $\frac{\partial H_t}{\partial W}$ as follows:

\begin{equation}
\frac{\partial H_t}{\partial W} = \sum\limits_{k=0}^{t} \frac{\partial H_t}{\partial \widehat{y}_t}\frac{\partial\widehat{y}_t}{\partial h_t} \left(\prod\limits_{j=k+1}^{t} \frac{\partial h_j}{\partial h_{j-1}}\right) \frac{\partial h_k}{\partial W}
\label{rnn_eq_gradient}
\end{equation}

where the inner product refers to the  chain rule until the lower limit of $k$. The only difference between this calculation and that of the standard back-propagation is that we sum up the gradients for $W$ at each time step. In an MLP, for example, since we don't share parameters across layers, so we don't need to add anything. The calculation above is the same when we calculate the gradient with respect to $U$ as well.

\begin{figure}[h]
\centering
\scalebox{.8}{
\begin{tikzpicture}

\begin{axis}[
    xmin=-2.5, xmax=2.5,
    ymin=-1.5, ymax=1.5,
    axis lines=center,
    axis on top=true,
    domain=-2.5:2.5,
    ylabel=$y$,
    xlabel=$x$,
    grid=both,
    grid style={line width=.1pt, draw=gray!10},
    legend pos=south east
    ]

    \addplot [mark=none,draw=red] {tanh(\x)};
    \addplot [blue, ultra thick] {(tanh(x+0.01)-tanh(x))/0.01};

    \draw [blue, dotted, thick] (axis cs:-2.5,-1)-- (axis cs:0,-1);
    \draw [blue, dotted, thick] (axis cs:+2.5,+1)-- (axis cs:0,+1);
    
    \addlegendentry{$y = tanh(x)$}
	\addlegendentry{$y = \frac{d(tanh(x))}{dx}$}
\end{axis}

\end{tikzpicture}
}
\caption{Activation function tanh and its derivative}
\label{derivative_tanh}
\end{figure}
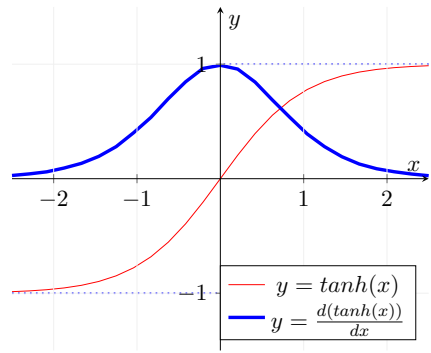

Equation \ref{rnn_eq_gradient} results in a square matrix including first-order partial derivatives of the vector of error function, also known as Jacobian matrix. One of the difficulties in the training process is the tendency of the derivatives of the activation function towards 0 as both extremes (the blue curve in figure \ref{derivative_tanh}) with a flat curve. Consequently, when applying the chain rule, the gradient of a layer tends towards 0 and vanishes completely after a few time steps. Vanishing gradients is a common problem, particularly in the networks that are intended to model a language. Long-distance dependencies are one of the characteristics of human language.

A more recent popular solution to confront the problem of the vanishing gradient is using Gated Recurrent Unit (GRU) or Long Short-Term Memory (LSTM) architectures. In our project, we used the latter, which is explained in the next section.

\subsubsection{Long short-term memory}

In the previous section, we stated that the vanishing gradient problem prevents standard RNNs from learning long-term dependencies. Long Short-Term Memory (LSTM) network is an architecture for the RNN, capable of learning long-term dependencies. They were initially introduced by Hochreiter and Schmidhube \cite{Hochreiter:1997:LSM:1246443.1246450} and were popularized by many other works in the following years (particularly in \cite{DBLP:journals/corr/Graves13,DBLP:journals/corr/abs-0705-2011,bayer2009evolving,graves2008supervised,graves2005framewise}). It is among the most widely used models in Deep Learning for NLP today. 

\begin{figure}[h]
  \centering
  \scalebox{.9}{
  \begin{tikzpicture}
  \coordinate (A) at (-2,0);
  \coordinate (B) at (6,0);  
  \coordinate (C) at (6,6);
  \coordinate (D) at (-2,6);
  \tikzstyle{neuron}=[circle,fill=red!25,minimum size=25pt,inner sep=0pt];
  \tikzstyle{gate}=[circle,fill=cyan!25,minimum size=25pt,inner sep=0pt];
  \tikzstyle{memory_cell}=[circle, thick, fill=green!35,minimum size=25pt,inner sep=0pt];
  \tikzstyle{multiplier}=[circle, thick, fill=black!25, minimum size=10pt,inner sep=0pt];
  \tikzstyle{nonlinearity}=[circle, thick, fill=blue!25,minimum size=15pt,inner sep=0pt];

  \node (input) at (-3, 5.5) [neuron]{$\vec{x}$};
  \draw[below left] (-3, 5) node {input};
  
  \node (output) at (8, 5.5) [neuron]{$\vec{h}$};
  \draw[above right] (7, 4.5) node {output};  
   
  \node (input_gate) at (0, 5.5) [gate]{$\vec{i}_t$};
  \node (output_gate) at (5.5, 5.5) [gate]{$\vec{o}_t$};
  \node (forget_gate) at (2, 0.5)[gate]{$\vec{f}_t$};
  
  \node (multiplier_1) at (0, 3.5)[multiplier]{$\times$};
  \node (multiplier_2) at (5.5, 3.5)[multiplier]{$\times$};
  \node (multiplier_3) at (2, 2)[multiplier]{$\times$};
  
  \node (memory) at (2, 3.5)[memory_cell]{$\vec{c}_t$};
  
  \node (nonlinearity_1) at (-1.5, 3.5)[nonlinearity]{$\tanh$};
  \node (nonlinearity_2) at (3.5, 3.5)[nonlinearity]{$\tanh$};
  
  \draw[thick] (A)--(B)--(C)--(D)--(A);

  \draw[->] (input) edge[-latex, out=0, in=180](input_gate);
  \draw[->] (input) edge[-latex, out=45, in=135](output_gate);
  \draw[->] (input) edge[-latex, out=-90, in=180](forget_gate);
  \draw[->] (input) edge[-latex, out=-45, in=90](nonlinearity_1);

  \draw[->] (multiplier_2) edge[-latex, out=0, in=180](output);
  
  \draw[->] (memory) edge[-latex](input_gate);
  
  \draw[->] (input_gate) edge[-latex](multiplier_1);
  \draw[->] (nonlinearity_1) edge[-latex](multiplier_1);
  
  \draw[->] (multiplier_1) edge[-latex](memory);
  \draw[->] (multiplier_3) edge[-latex, out=45,in=-45] (memory);
  
  \draw[->] (memory) edge[-latex, out=-135,in=135] (multiplier_3);
  \draw[->] (forget_gate) edge[-latex](multiplier_3);
  
  \draw[->] (memory) edge[-latex, out=-145,in=135](forget_gate);  

  \draw[->] (memory) edge[-latex](nonlinearity_2);
  
  \draw[->] (output_gate) edge[-latex](multiplier_2);
  \draw[->] (nonlinearity_2) edge[-latex](multiplier_2);  

  \draw[->] (memory) edge[-latex] (output_gate);

\begin{customlegend}[legend cell align=right, 
legend entries={ 
weighted connection,
activation function,
cell state (memory), 
gates,
input and state units,
LSTM
},
legend style={at={(10.2,2.7)},font=\footnotesize}] 
    \addlegendimage{->,black,opacity=0.4}
    \addlegendimage{mark=ball,color=blue!25,draw=white}
    \addlegendimage{mark=ball,ball color=green!35,draw=white}
    \addlegendimage{mark=ball,ball color=cyan!25,draw=white}
    \addlegendimage{mark=ball,ball color=red!25,draw=white}
    \addlegendimage{area legend,black,fill=white}
\end{customlegend}	
	
\end{tikzpicture}
}
\caption{Architecture of LSTM. The gates are implemented to compute a value between 0 and 1 and are multiplied to partially allow or deny information to flow into or out of the memory (cell state). The final value for a given input is then transfered to the output connection of the layer.}
\label{lstm}
\end{figure}
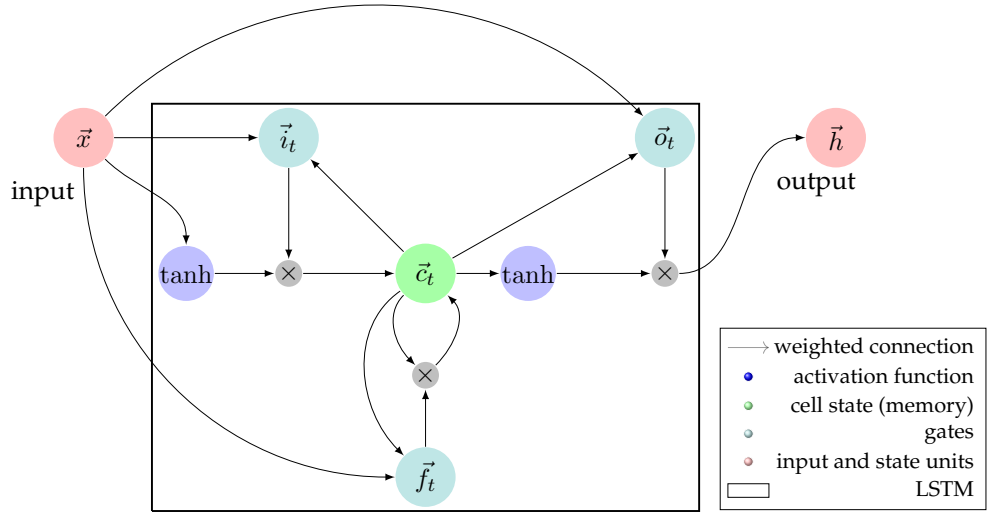

An LSTM network is composed of memory cells and gate units to calculate the internal states. Three gates control the behavior of the memory cells: ignore gate  $i$, forget gate $f$ and output gate $o$. These gates intuitively determine precisely how much of a given vector is to be taken into account at a specific time step $t$. For instance, the input gate defines how much of current input should go through, the forget input defines much of the previous state pass and the output gate indicates how much of the internal state be influenced in the next time steps and next layers. The full architecture of the LSTM is illustrated in figure \ref{lstm}. Formally, an LSTM is defined as:

\begin{center}
\begin{gather}
i_t= \sigma(W_{xi}x_t+W_{hi}h_{t-1}+W_{ci}c_{t-1}+bi)\\
f_t=\sigma(W_{xf}x_t+W_{hf}h_{t-1}+W_{cf}c_{t-1}+b_f) \\
o_t=\sigma(W_{xo}x_t+W_{ho}h_{t-1}+W_{co}c_t+b_o)\\
c_t=f_t.c_{t-1}+i_t.tanh(W_{xc}x_t+W_{hc}h_{t-1}+b_c)\\
h_t=o_ttanh(c_t)
\end{gather}
\end{center}

where $i_t$, $f_t$ and $o_t$ respectively refer to the input, forget and output gates at time step $t$. $c_t$ is the memory cell vector at time step $t$ and $h_t$ is the output layer. Note that the gates have the same equations, but with different parameter matrices. Various variations of LSTM can be found in \cite{greff2016lstm}.

\subsection{Bidirectional Recurrent Neural Network}

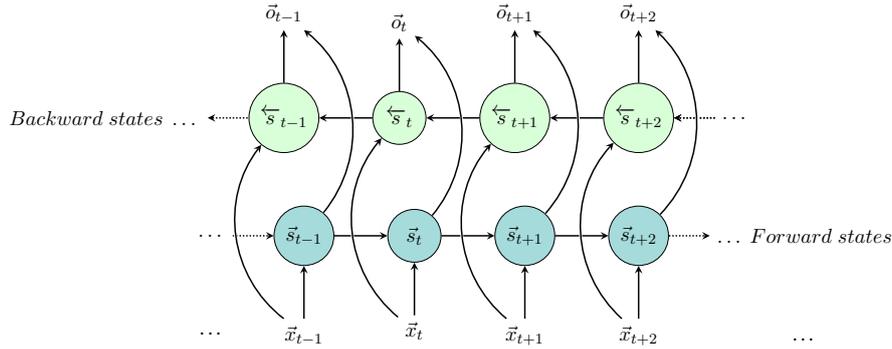
\begin{figure}[h]
\centering
\scalebox{.7}{
\begin{tikzpicture}
	\node[rectangle] (Y0) at (0, 0) {$\dots$};
	\node[circle, draw, right=2em of Y0, minimum height=1cm, minimum width=1cm,fill=cyan!35] (RNN) {$\vec{s}_{t-1}$};
	\node[circle, right=of RNN, draw, minimum height=1cm, minimum width=1cm,fill=cyan!35] (RNN2) {$\vec{s}_{t}$};
	\node[circle, right=of RNN2, draw, minimum height=1cm, minimum width=1cm,fill=cyan!35] (RNN3) {$\vec{s}_{t+1}$};
			
	\node[circle, right= of RNN3, draw, minimum height=1cm, minimum width=1cm,fill=cyan!35] (RNN4) {$\vec{s}_{t+2}$};
	\node[rectangle, right=2em of RNN4] (RNN5) {$\dots\;Forward\;states$};
			
	\node[circle, above=of RNN4, draw, minimum height=1cm, minimum width=1cm,fill=green!15] (R25) {$\overleftarrow{s}_{t+2}$};
	\node[circle, left=of R25, minimum height=1cm, minimum width=1cm, draw,fill=green!15] (R24) {$\overleftarrow{s}_{t+1}$};
	\node[circle, left=of R24, draw, minimum height=1cm, minimum width=1cm,fill=green!15] (R23) {$\overleftarrow{s}_{t}$};
	\node[circle, left=of R23, draw, minimum height=1cm, minimum width=1cm,fill=green!15] (R22) {$\overleftarrow{s}_{t-1}$};
	\node[circle, left=2em of R22] (R21) {$Backward\;states\;\dots$};
	\node[right=2em of R25] (Y20) {$\dots$};
			
	\node[below=of RNN] (X1) {$\vec{x}_{t-1}$};
	\node[below=of RNN2] (X2) {$\vec{x}_{t}$};
	\node[below=of RNN3] (X3) {$\vec{x}_{t+1}$};
	\node[below=of RNN4] (X4) {$\vec{x}_{t+2}$};
	\node[above=of R25] (Y5) {$\vec{o}_{t+2}$};
	\node[above=of R24] (Y4) {$\vec{o}_{t+1}$};
	\node[above=of R23] (Y3) {$\vec{o}_{t}$};
	\node[above=of R22] (Y2) {$\vec{o}_{t-1}$};
			
	\draw[-stealth, thick] (X1) -- (RNN);
	\draw[-stealth, thick] (X2) -- (RNN2);
	\draw[-stealth, thick] (X3) -- (RNN3);
	\draw[-stealth, thick] (X4) -- (RNN4);
	\draw[-stealth, thick, densely dotted] (Y0) -- (RNN);
	\draw[-stealth, thick] (RNN) -- node[above, pos=0.35] {} (RNN2);
	\draw[-stealth, thick] (RNN2) -- node[above, pos=0.35] {} (RNN3);
	\draw[-stealth, thick] (RNN3) -- node[above, pos=0.35] {} (RNN4);
	\draw[-stealth, densely dotted, thick] (RNN4) -- (RNN5);
	\node[below=4em of Y0] (d) {\dots};
	\node[below=4em of RNN5] (d) {\dots};
			
	\path[-stealth, ultra thick, white] (X1) edge[bend left=45] (R22);
	\path[-stealth, thick] (X1) edge[bend left=45] (R22);
	\path[-stealth, ultra thick, white] (X2) edge[bend left=45] (R23);
	\path[-stealth, thick] (X2) edge[bend left=45] (R23);
	\path[-stealth, ultra thick, white] (X3) edge[bend left=45] (R24);
	\path[-stealth, thick] (X3) edge[bend left=45] (R24);
	\path[-stealth, ultra thick, white] (X4) edge[bend left=45] (R25);
	\path[-stealth, thick] (X4) edge[bend left=45] (R25);
	\draw[-stealth, densely dotted, thick] (Y20) -- (R25);
			
	\draw[-stealth, thick] (R22) -- (Y2);
	\draw[-stealth, thick] (R23) -- (Y3);
	\draw[-stealth, thick] (R24) -- (Y4);
	\draw[-stealth, thick] (R25) -- (Y5);
		
	\draw[stealth-, densely dotted, thick] (R21) -- (R22);
	\draw[stealth-, thick] (R22) -- node[above, pos=0.65] {} (R23);
	\draw[stealth-, thick] (R23) -- node[above, pos=0.65] {} (R24);
	\draw[stealth-, thick] (R24) -- node[above, pos=0.65] {} (R25);
	\draw[-stealth, densely dotted, thick] (Y20) -- (R25);	
			
	\path[-stealth, ultra thick, white] (RNN) edge[bend right=45] (Y2);
	\path[-stealth, thick] (RNN) edge[bend right=45] (Y2);
	\path[-stealth, ultra thick, white] (RNN2) edge[bend right=45] (Y3);
	\path[-stealth, thick] (RNN2) edge[bend right=45] (Y3);
	\path[-stealth, ultra thick, white] (RNN3) edge[bend right=45] (Y4);
	\path[-stealth, thick] (RNN3) edge[bend right=45] (Y4);
	\path[-stealth, ultra thick, white] (RNN4) edge[bend right=45] (Y5);
	\path[-stealth, thick] (RNN4) edge[bend right=45] (Y5);
			
\end{tikzpicture}
}
\caption{A bidirectional recurrent neural network model}
\label{fig_bidirectional}
\end{figure}

So far, we have focused on the RNNs that consider the past observation in order to predict the next symbol in a sequence. For many sequence-labeling tasks, e.g., predicting the correct character for a given input, in addition to the past context, we would like to have access to the future context. To do so, we can use two RNN models; one that reads through the input sequence forwards (left-to-right propagation) and the other backwards (right-to-left propagation), both with two different hidden units but connected to the same output(figure \ref{fig_bidirectional}). This model is called Bidirectional Recurrent Neural Network (BRNN) \cite{schuster1997bidirectional}.

The right-to-left propagation for the BRNN hidden layers is the same as for a simple RNN, except that the input sequence is presented in the opposite direction to the two hidden layers. Processing the hidden layers of the two RNN models is necessary to update the output layer. The following equation shows the mathematical formulation behind setting up the bidirectional RNN hidden layer. Note that calculating the hidden layers follows the same formulation mentioned in \ref{rnn_equation_1}:

\begin{gather}
\vec{h_t}=tanh(\vec{W}x_t+\vec{U}\vec{h}_{t-1}+\vec{b})\\
\overleftarrow{h_t}=tanh(\overleftarrow{W}x_t+\overleftarrow{U}\overleftarrow{h}_{t+1}+\overleftarrow{b})
\end{gather}

The corresponding calculation for each direction is shown by using vector on each formulation. These two calculations are concatenated as $h_t$ in equation \ref{rnn_equation_1}. In order to train such a model, we will need two twice as much memory for the parameters.

BRNNs have previously given improved results in various domains in speech processing \cite{schuster1999,SCJ:SCJ3}. More recently, deep BRRN have been used where each lower layer  feeds  the  next  layer \cite{mohamed2015deep,graves2013hybrid}.

\subsection{Sequence-to-sequence models}

Apart from long dependency, another challenge in language models using neural networks is the variable-length output spaces, e.g., words and sentences. Sequence-to-sequence neural models, or Encoder-Decoder network, have demonstrated the ability to address this challenge of the variable length of input and output. It is a generative neural network model that, given a string of inputs, produces a string of outputs, both of arbitrary lengths.  

\begin{figure}[h]
\centering
\scalebox{.7}{
\begin{tikzpicture}[
  hid/.style 2 args={rectangle split, rectangle split horizontal, draw=#2, rectangle split parts=#1, outer sep=1mm},
  label/.style={font=\small}]
    \tikzset{>=stealth',every on chain/.append style={join},
      every join/.style={->}}  

  \foreach \step in {1,2,3}
    \node[hid={4}{black},fill=violet!10] (h\step) at (3*\step, 2) {};

  \foreach \step in {1,2,3}
    \node[hid={4}{black},fill=yellow!10] (w\step) at (3*\step, 0) {};    
    
  \foreach \step in {1,2,3}
    \draw[->] (w\step.north) -> (h\step.south);

  \foreach \step/\next in {1/2,2/3}
    \draw[->] (h\step.east) -> (h\next.west);
  \foreach \step in {4,5,6}
    \node[hid={4}{black},fill=red!20] (hd\step) at (3*\step, 3.5) {};

  
  \foreach \step in {4,5,6}
    \node[hid={4}{black},fill=green!20] (softmax\step) at (3*\step, 4.8) {};    
    

  \foreach \step/\next in {4/5,5/6}
    \draw[->] (hd\step.east) -> (hd\next.west);
    
	
   \node[] (h0) at (0.6, 2) {};
   \node[] (h4) at (11.2, 2) {};
    \draw[->] (h0.east) -> (h1.west);
 
  
  \draw[->] (h3.east) to [->,out=40,in=150] (hd4.west);
  
  \draw[->] (12,3.7) -> (12,4.5);
  \draw[->] (15,3.7) -> (15,4.5);
  \draw[->] (18,3.7) -> (18,4.5);
  
  \draw[->] (3,-1.3) -> (3,-.2);
  \draw[->] (6,-1.3) -> (6,-.2);
  \draw[->] (9,-1.3) -> (9,-.2);
  
  \draw[->] (12,4.8) -> (12,6.2);
  \draw[->] (15,4.8) -> (15,6.2);
  \draw[->] (18,4.8) -> (18,6.2);
  \draw[->] (19,3.5) -> (20,3.5);
  
  \draw [thick, -latex] (softmax4.north) to [->,out=40,in=255] (hd5.south);
    \draw [thick, -latex] (softmax5.north) to [->,out=40,in=255] (hd6.south);

    
 \node[label] (hlabel3) at (13.2, 1.2){$\mathbf{h}_3 = \mathrm{tanh}(Wx_2+U\mathbf{h}_2+b )$};
  \node[label] at (7.3, 2.2){$h_2$};
  \node[label] at (4.3, 2.2){$h_1$};

  \foreach \step in {1,2,3}
    \node[label] (wlabel\step) at (3*\step-1.2, 1) {$embedding\;x_\step$};
    
    
   \foreach \step in {1,2,3}
    \node[label] (wlabel\step) at (3*\step, -1.5) {$x_\step$};
        
    \foreach \step in {4,5,6}
    \node[label] (wlabel\step) at (3*\step-.8, 5.5) {$softmax$};
    
    \foreach \step in {4,5,6}
    \pgfmathsetmacro\result{\step-3}
    \node[label] (wlabel\step) at (3*\step, 6.5) {$o_{\pgfmathprintnumber{\result}}$};
    
    \foreach \step in {4,5,6}
    \pgfmathsetmacro\result{\step-3}
    \node[label] (wlabel\step) at (3*\step+1.5,3.2) {$s_{\pgfmathprintnumber{\result}}$};
\end{tikzpicture}
}
\caption{Sequence-to-sequence model for three time steps. The state vector of the last layer of the encoder is passed to the decoder as input. Consequently, each output of the decoder is used as the input of the succeeding layer in the decoder.}
\label{fig_encoder_decoder_softmax}
\end{figure}
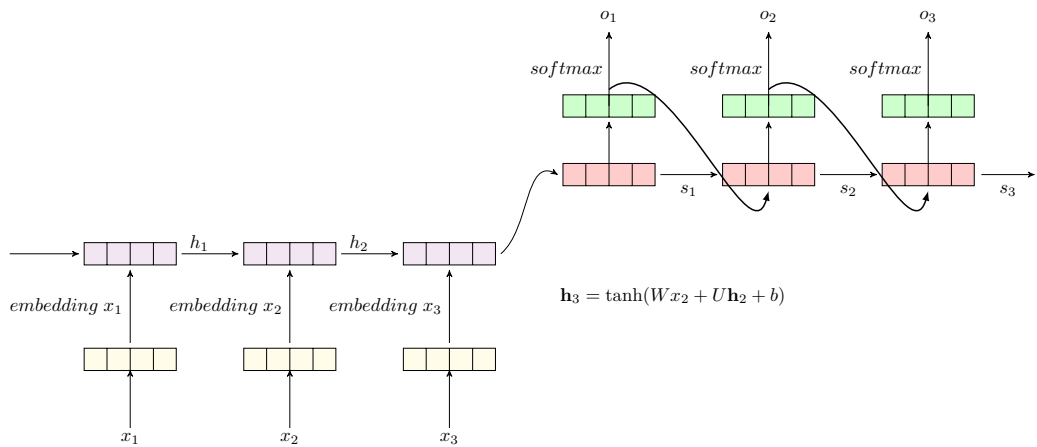

The sequence-to-sequence model is composed of two processes : \emph{encoding} and \emph{decoding}. In the encoding process, the input sequence $x=(x_1,x_2,...,x_T)$ is fed into the encoder, which is basically an RNN model. Then, unlike a simple RNN where the output of each state unit is taken into account, only the hidden unit of the last layer $h_t$ is kept; this vector, often called \emph{sentence embedding} or \emph{context vector} $c$, is intended to contain a representation of the input sentence:

\begin{gather}
h_t=LSTM(x_t,h_{t-1})\\
c = tanh(h_{T})
\end{gather}

where $h_t$ is a hidden state at time $t$, and c is the context vector of the hidden layers of the encoder.

On the other hand, by passing the context vector $c$ and all the previously predicted words $\{y_1,y_2,...,y_{'{t}-1}\}$ to the decoder, the decoding process predicts the next word $'{y_t}$. In other words, the decoder defines a probability over the output $y$ by decomposing the joint probability as follows:

\begin{gather}
p(y) = \prod_{t=1}^T p(y_t|\{y_1,y_2,...,y_{t-1}\},c)\\
p(y_t|\{y_1,y_2,...,y_{t-1}\},c) = tanh(y_{t-1}, h_t,c)
\end{gather}

where $y=(y_1,y_2,...,y_{T})$ and $h_t$ is the hidden unit.

Figure \ref{fig_encoder_decoder_softmax} depicts an encoding and a decoding process the input sequence of $x_1, x_2,x_3$. Softmax function is used in order to normalize the probability distribution for the output. Later, the Softmax output would be used for calculating the error loss as well.  

In some of the first studies that used encoder-decoder RNN models, a fixed-length vector is used to represent the context of a source sentence \cite{cho2014learning,sutskever2014sequence}. The fact that a fixed-length, regardless of the size of the input, is used, means a limitation on the model to represent variable-size input strings. More recently, Bahdanau et al. \cite{bahdanau2014neural} explored the attention mechanism in NMT which is discussed in the following section.

\subsection{Attention mechanism}

Attention mechanism was introduced to address the limitation of modeling long dependencies and the efficient usage of memory for computation. The attention mechanism intervenes as an intermediate layer between the encoder and the decoder, having the  objective  of capturing  the  information  from  the  sequence of tokens  that  are  relevant  to  the contents of the sentence \cite{bahdanau2014neural}.

\begin{figure}[h]
\centering
\scalebox{.7}{
\begin{tikzpicture}[
  hid/.style 2 args={rectangle split, rectangle split horizontal, draw=#2, rectangle split parts=#1, outer sep=1mm},
  label/.style={font=\small}]
    \tikzset{>=stealth',every on chain/.append style={join},
      every join/.style={->}}  

  \foreach \step in {1,2,3}
    \node[hid={4}{black},fill=violet!10] (h\step) at (3*\step, 2) {};

  \foreach \step in {1,2,3}
    \node[hid={4}{black},fill=yellow!10] (w\step) at (3*\step, 0) {};    
    
  \foreach \step in {1,2,3}
    \draw[->] (w\step.north) -> (h\step.south);

  \foreach \step/\next in {1/2,2/3}
    \draw[->] (h\step.east) -> (h\next.west);
  \foreach \step in {4,5,6}
    \node[hid={4}{black},fill=red!20] (hd\step) at (3*\step, 3.5) {};

   \node[hid={1}{black},fill=white] (attention) at (2*3, 3.5) {$attention$};

  \foreach \step in {4,5,6}
    \node[hid={4}{black},fill=green!20] (softmax\step) at (3*\step, 4.8) {};    
    

  \foreach \step/\next in {4/5,5/6}
    \draw[->] (hd\step.east) -> (hd\next.west);
    
  
   \foreach \step in {1,2,3}
    \draw[->] (h\step.north) -> (attention.south);
    
    \draw[dashed,draw=red] (attention.north) to [->,bend left] (9,3.5);
    
       \foreach \step in {4,5,6}
    \draw[dashed,->,draw=red] (9,3.5) to [->,out=-180,in=-50] (hd\step.south);
    
	
   \node[] (h0) at (0.6, 2) {};
   \node[] (h4) at (11.2, 2) {};
    \draw[->] (h0.east) -> (h1.west);
 
  
  \draw[->] (h3.east) to [->,out=40,in=150] (hd4.west);
  
  \draw[->] (12,3.7) -> (12,4.5);
  \draw[->] (15,3.7) -> (15,4.5);
  \draw[->] (18,3.7) -> (18,4.5);
  
  \draw[->] (3,-1.3) -> (3,-.2);
  \draw[->] (6,-1.3) -> (6,-.2);
  \draw[->] (9,-1.3) -> (9,-.2);
  
  \draw[->] (12,4.8) -> (12,6.2);
  \draw[->] (15,4.8) -> (15,6.2);
  \draw[->] (18,4.8) -> (18,6.2);
  \draw[->] (19,3.5) -> (20,3.5);
  
  \draw [thick, -latex] (softmax4.north) to [->,out=40,in=255] (hd5.south);
    \draw [thick, -latex] (softmax5.north) to [->,out=40,in=255] (hd6.south);

    
 \node[label] (hlabel3) at (13.2, 1.2){$\mathbf{h}_3 = \mathrm{tanh}(Wx_2+U\mathbf{h}_2+b )$};
  \node[label] at (7.3, 2.2){$h_2$};
  \node[label] at (4.3, 2.2){$h_1$};

  \foreach \step in {1,2,3}
    \node[label] (wlabel\step) at (3*\step-1.2, 1) {$embedding\;x_\step$};
    
    
   \foreach \step in {1,2,3}
    \node[label] (wlabel\step) at (3*\step, -1.5) {$x_\step$};
        
    \foreach \step in {4,5,6}
    \node[label] (wlabel\step) at (3*\step-.8, 5.5) {$softmax$};
    
    \foreach \step in {4,5,6}
    \pgfmathsetmacro\result{\step-3}
    \node[label] (wlabel\step) at (3*\step, 6.5) {$o_{\pgfmathprintnumber{\result}}$};
    
    \foreach \step in {4,5,6}
    \pgfmathsetmacro\result{\step-3}
    \node[label] (wlabel\step) at (3*\step+1.5,3.2) {$s_{\pgfmathprintnumber{\result}}$};
\end{tikzpicture}
}
\caption{Attention-based sequence-to-sequence model for three time steps. The red arrows depict the connection of the intermediate vector $c$ with the state units of the decoder.}
\label{fig_attention_model}
\end{figure}
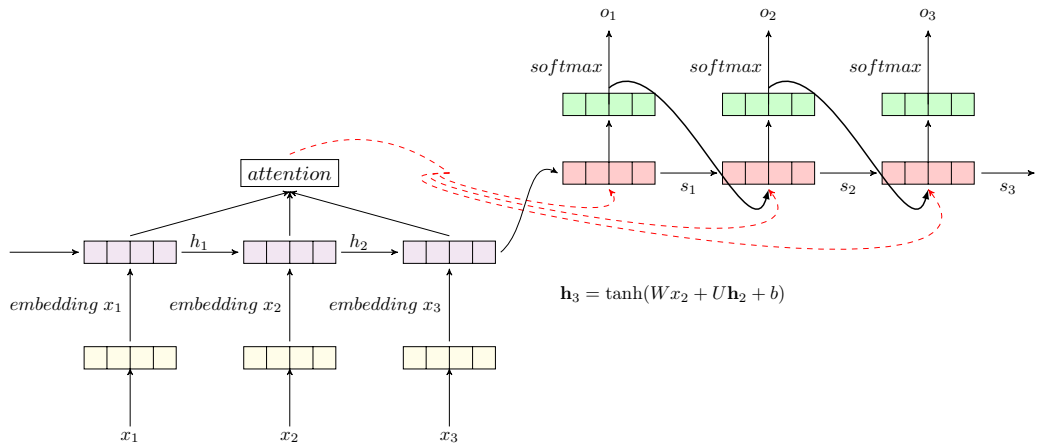

In an attention-based model, a set of attention weights is first calculated. These are multiplied by the encoder output vectors to create a weighted combination. The result should contain information about that specific part of the input sequence, and thus help the decoder select the correct output symbol. Therefore, the decoder network can use different portions of the encoder sequence as context, while it is processing the decoder sequence. This justifies the variable-size representation of the input sequence. This mechanism is shown in figure \ref{fig_attention_model}.

Unlike the encoder-decoder model that uses the same context vector for every hidden state of the decoder, the attention mechanism calculates a new vector $c_t$ for the output word $y_t$ at the decoding step $t$. It can be defined as:

\begin{equation}
c_t=\sum_{j=1}^{T}a_{tj}h_j
\end{equation}

where $h_j$ is the hidden state of the word $x_j$, and $a_{tj}$ is the weight of $h_j$ for predicting $y_t$. This vector is also called \emph{attention vector} wish is generally calculated with a softmax function:

\begin{gather}
\alpha_{ij}=\frac{exp(e_{ij})}{\sum_{k=1}^T exp(e_{ik})} \\
e_{ij} = attentionScore(s_{i-1},h_j)
\end{gather}

where the \texttt{attentionScore}, as an arbitrary function in the calculation, which scores the quality of alignment between input $j$ and output $i$. It is parameterized as a feed-forward neural network in the original paper \cite{bahdanau2014neural}. This vector is then normalized into the actual attention vector by usng a softmax function over the scores:

\begin{equation}
\alpha_t = softmax(attentionScore)_t
\end{equation}

This attention vector is then used to weight the encoded representation of the hidden state $h_j$. Having said that in addition to this formulation, some others have been proposed for attention, particularly in \cite{luong2015effective}.

\section{Evaluation metrics}
\label{evaluation_metrics}

This section presents evaluation methods that would be used to evaluate the models. Although various methods can be used to evaluate a machine translation system, in the case of machine correction, evaluation may be limited to a binary classification of predictions where the matching elements are considered as true predictions and the others, incorrect predictions. However, a good evaluation must include more details about this comparison. 

\begin{testexample}[Comparison of two error correction systems\label{exe:comparison_2_correctors}]
\begin{table}[H]
\centering
\begin{tabular}{l|l}
\hline
\multicolumn{2}{l}{\textbf{Input}: The burgening cherry trees are a signe that spring ishere.}                              \\ 
\multicolumn{2}{l}{\textbf{Gold-standard}: \{burgening $\rightarrow $ burgeoning, signe $\rightarrow$ sign, ishere $\rightarrow$ is here\} }                      \\ \hline\hline
\multicolumn{2}{l}{Hypothesis A: \{burgening $\rightarrow $ burning, a signe $\rightarrow$ assigned\} }                      \\ \hline
\multicolumn{2}{l}{Hypothesis B: \{burgening $\rightarrow $ burgeoning, signe $\rightarrow$ sign, ishere.$\rightarrow$is here., . $\rightarrow\emptyset$ \} }                      \\ \hline

\end{tabular}
\end{table}
\end{testexample}

Example \ref{exe:comparison_2_correctors} illustrates a comparison between the results of two different correction systems. For the given input phrase, two hypotheses are proposed by the two correction systems: Hypothesis A yields, "\textit{The burning cherry trees are assigned that spring ishere.}", while hypothesis B produces the same expected phrase of the gold-standard annotation. 

How would one evaluate the performance of these correction systems? The output of system A and B includes respectively 9 and 12 tokens. How can we compare the elements of the predicted and gold-standard phrases pairwise, while they have different lengths?  A further challenge arises when we have different edit actions that yield the same output which is the case of correction system B. Finally, how can we distinguish between two incorrect predictions and determine their quality? For instance, correction system A predicts "\textit{burning}" as a correction for "\textit{burgening}", which is not true, but its prediction includes similar letters that indicate a better performance in comparison to a completely non-related correction. 

Over the years, a number of metrics have been proposed for error correction systems evaluation, each motivated by weaknesses of previous metrics. There is no single best evaluation metric and the performance of a metric depends on the research goals and application \cite{chodorow2012problems}. Thus, we have evaluated our system based on the most popular metrics to date.

\subsection{Standard metrics}
\label{sec_stan_metrics}


In classification tasks, \emph{accuracy} is one of the most used performance measures. This metric corresponds to the ratio of correctly classified inputs to the total number of inputs. More formally, accuracy can be defined as:  

\begin{equation}
acc=\frac{TP+TN}{TP+TN+FP+FN}
\end{equation}

where TN (true negative) corresponds to the number of false inputs that were correct classified, TP (true positive) corresponds to the number of positive inputs that were classified correctly, FN (false negative) corresponds to the number of positive samples incorrectly classified as positive, and finally, FP (false positive) corresponds to the number of negative inputs incorrectly classified as positive. 

One drawback of this metric is that correction actions are completely ignored. Consider correction system B in example \ref{exe:comparison_2_correctors} which yields the expected output but with different correction actions. Even if $accuracy=1$ in this case, the performance of actions are not reflected. 

In order to take the correction actions into account, we define two additional performance metrics, \emph{precision} and \emph{recall}.  Given a set of $N$ sentences, where $G_i$ is the set of gold-standard edits for sentence $i$, and $P_i$ is the set of predicted edits for sentence $i$, we define precision, recall and F-measure as follows:

\begin{equation}
precision =\frac{TP}{TP+FP} = \frac{\sum_{i=1}^N | G_i \cap P_i|}{\sum_{i=1}^N |P_i|}
\label{eq_precision}
\end{equation}

\begin{equation}
recall = \frac{TP}{TP+FN}= \frac{\sum_{i=1}^N | G_i \cap P_i|}{\sum_{i=1}^N |G_i|} 
\label{eq_recall}
\end{equation} 

where the intersection between $G_i$ and $P_i$ indicates the matches between the two sets. A measure that combines $precision$ and $recall$ is the harmonic mean of them, also known as $F_\beta-measure$. It is defined as:

\begin{equation}
F_\beta = (1 + \beta^2). \frac{precision \times recall}{\beta ^2 \times precision + recall} 
\label{eq_fmeasure}
\end{equation} 

Depending on the usage, different values are attributed to $\beta$. The most common case is in $F_{\beta=1}$, also known as \textit{F-score}, which weights recall and precision evenly. As the evaluation metric in our research, we have used $F_{0.5}$, since it places twice as much emphasis on precision than recall, while $F_1$ weighs precision and recall equally. This metric has also been used in CoNLL-2014 shared task \cite{ng2014conll}. Our $F_{0.5}$ is defined as follows:

\begin{equation}
F_{0.5} = (1 + {0.5}^2). \frac{precision \times recall}{{0.5} ^2 \times precision + recall} 
\end{equation}

To illustrate, consider correction system B in example \ref{exe:comparison_2_correctors}. Based on our defined metrics, the performance of this correction is: $P=2/4$, $R=2/3$ and $F_{0.5}=(1+0.5^2)\times P \times R / ({0.5^2 \times P + R}) = 10/19$. Although this correction system yields the expected phrase of the gold-standard annotation in the output, these metrics still do not provide sufficient information. Our correction system produces its predicted text for the given source phrase, and we should provide correction actions to be compared with the gold-standard annotations.

In the case of evaluation of machine correction systems, $P$, $R$ and $F_{0.5}$ do not provide an indicator of improvement on the original text. Given this, we opt to use more ameliorated metrics such as \textit{MaxMatch ($M^2$)}, \emph{I-measure},  \emph{BLEU} and \emph{GLEU}, thus gaining better insight into system performance. Each metric is explained in the following sections.

\subsection{MaxMatch ($M^2$)}
\label{m2_method}

Dahlmeier et al. \cite{dahlmeier2012better} proposed a more accurate evaluation method, which is known as MaxMatch($M^2$), for efficiently computing the sequence of phrase-level edits between a source text and the predicted text. The main idea is to determine the set of edits that most often matches the gold-standard. Thus, the algorithm yields the set of phrase-level edits with the maximum overlap with the gold-standard. The edits are, then, scored using $F_\beta$ measure.

\subsubsection{Method}

Given a set of source sentences $S=\{s_1,...,s_n\}$ with a set of gold-standard annotation $G=\{g_1,...,g_n\}$, our error correction system yields a set of predictions $P=\{p_1,...,p_n\}$. Each annotation $g_i$ contains a set of edits $g_i=\{g_i^1,...,g_i^m\}$. $M^2$ method, in the first step, extracts a set of system edits $e_i$ for each pair of ($s_i$, $p_i$). Thus, an edit lattice structure is constructed from a source-prediction pair. Then, the optimal sequence of edits is determined by solving a shortest path search through the lattice. 

\paragraph{Constructing edit lattice}
Assuming that we have tokenized sentence $i$ in the source of sentences $S$ as $s_i=\{s_i^1,...,s_i^r\}$ for which its relevant prediction is $p_i=\{p_i^1,...,p_i^r\}$. We start by calculating the Levenshtein distance in pair ($s_i$, $h_i$). In the resulting edit distance matrix, the last element represents the edit distance value for the two sets. By determining the shortest paths from the first element of the matrix([0,0]) to the last element, we can construct a lattice structure in which each vertex corresponds to a cell in the matrix and each edge to an atomic edit operations. Atomic edit operation in the Levenshtein algorithm involves deleting a token, inserting a token, substituting a token or leaving a token unchanged.

\begin{testexample}[]
\\ \textbf{Input}: the greater the Levenshtein distances , more different strings are .
\\ \textbf{Prediction}: the greater the Levenshtein distances , the more different strings are .
\\ $G=$\{distances $\rightarrow$ distance, more $\rightarrow$ the more\}
\\ $P=$\{more $\rightarrow$ the more\} 
\label{example_levenshtein}
\end{testexample}

Example \ref{example_levenshtein} presents a phrase in input and the prediction of the correction system. Figure \ref{levenshteinMatrix} indicates the Levenshtein distance matrix between the input and the prediction. The shortest path that leads to the minimum edit distance in the bottom-right corner of the matrix is identified by circles. Therefore, by applying a breadth-first search on the matrix, we construct the corresponding lattice.

\begin{figure}[h]
\centering
\scalebox{.7}{	
\begin{tabular}{|l||l|l|l|l|l|l|l|l|l|l|l|l|l|}
\hline
            &    & the & greater & the & Levenshtein & distances  & , & the & more & different & strings   & are    & .	 \\ \hline \hline
            & \Circled{\textbf{0}}  & 1      & 2    & 3   & 4 	 & 5 	 & 6  & 7  	   & 8 & 9 	& 10 & 11 & 12 \\ \hline
the     	& 1  &\Circled{\textbf{0}} &1 &2 &3 &4 &5 &6 &7 &8 &9 &10 &11     \\ \hline
greater     & 2  &1 &\Circled{\textbf{0}} &1 &2 &3 &4 &5 &6 &7 &8 &9 &10     \\ \hline
the         & 3  &2 &1 &\Circled{\textbf{0}} &1 &2 &3 &4 &5 &6 &7 &8 &9     \\ \hline
Levenshtein & 4  &3 &2 &1 &\Circled{\textbf{0}} &1 &2 &3 &4 &5 &6 &7 &8 \\ \hline
distances   & 5  &4 &3 &2 &1 &\Circled{\textbf{0}} &1 &2 &3 &4 &5 &6 &7      \\ \hline
,         	& 6  &5 &4 &3 &2 &1 &\Circled{\textbf{0}} &\Circled{\textbf{1}} &2 &3 &4 &5 &6     \\ \hline
more        & 7  &6 &5 &4 &3 &2 &1 &1 &\Circled{\textbf{1}} &2 &3 &4 &5      \\ \hline
different 	& 8  &7 & 6 &5 &4 &3 &2 &2& 2& \Circled{\textbf{1}} &2 &3 &4     \\ \hline
strings     & 9  &8 &7 &6 &5 & 4 &3 &3& 3& 2& \Circled{\textbf{1}} &2 &3      \\ \hline
are         & 10 &9 &8 &7 &6 &5 &4 &4 &4 &3 &2 &\Circled{\textbf{1}} &2     \\ \hline
.         	& 11 &10 &9 &8 &7 &6 &5 &5 &5 &4 &3 &2 &\Circled{\textbf{1}}     \\ \hline

\end{tabular}
}
\caption{Levenshtein edit distance matrix for the input and the prediction in example \ref{example_levenshtein}}
\label{levenshteinMatrix}
\end{figure}

\begin{figure}[h]
\centering
\scalebox{.7}{
\begin {tikzpicture}[-latex ,auto ,node distance =4 cm and 5cm ,on grid ,
semithick ,
state/.style ={ circle,fill=blue!20}]
\node[state] (A) at (0, 11) {0,0};
\node[state] (B) at (2, 11) {1,1};
\node[state] (C) at (4.5, 11) {2,2};
\node[state] (D) at (6.5,11) {3,3};
\node[state] (E) at (10, 11) {4,4};
\node[state] (F) at (6.5,8.5){5,5};
\node[state] (G) at (6.5,4.5) {6,6};
\node[state] (H) at (10,8.5) {6,7};
\node[state] (I) at (10,4.5) {7,8};
\node[state] (J) at (6.5,2) {8,9};
\node[state] (K) at (4,2) {9,10};
\node[state] (L) at (2,2) {10,11};
\node[state] (O) at (0,2) {11,12};

\path (A) edge [->] node[above] {the (1)} (B);
\path (B) edge [->] node[above] {greater (1)} (C);
\path (C) edge [->] node[above] {the (1)} (D);
\path (D) edge [->] node[above] {Levenshtein (1)} (E);
\path (E) edge [bend left =0] node[above left] {distances (1)} (F);
\path (F) edge [->] node[left] {, (1)} (G);
\path (G) edge [bend right =35] node[left =0.15 cm] {$\emptyset$/the (1)} (H);
\path (H) edge [->] node[right] {more (1)} (I);
\path (I) edge [bend left =35] node[right=0.15 cm] {different (1)} (J);
\path (J) edge [->] node[below =0.15 cm] {strings (1)} (K);
\path (K) edge [->] node[below =0.15 cm] {are (1)} (L);
\path (L) edge [->] node[below =0.15 cm] {. (1)} (O);

\path (E) edge [->] node[right] {distance, / distance, the (3)} (H);
\path (F) edge [->] node[below = 0.15 cm] {, / , the (2)} (H);
\path (G) edge [->] node[below] {{\small more / the more (\textbf{-48})}} (I);
\path (G) edge [->] node[left] {more different / the more different (3)} (J);
\end{tikzpicture}
}
\caption{Edit lattice of matrix \ref{levenshteinMatrix}. Transitive edits are weighted by the sum of the consisting parts. }
\label{latticeExample}
\end{figure}
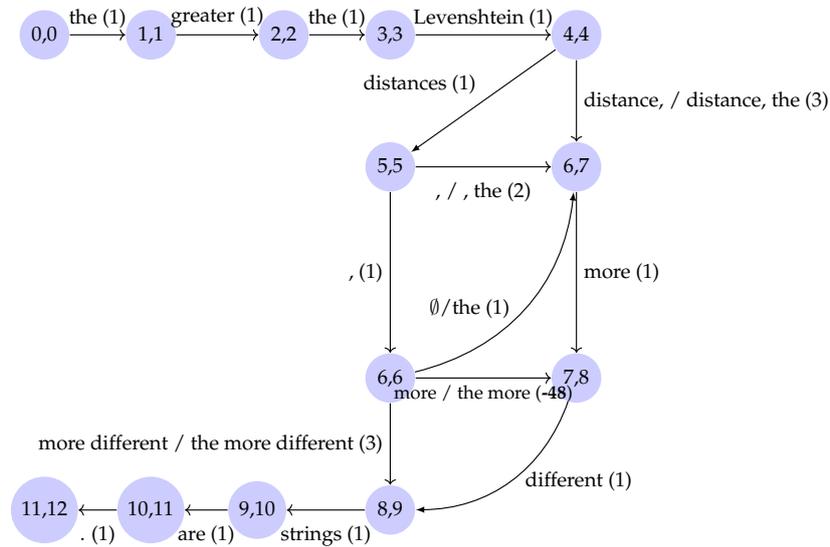

Figure \ref{latticeExample} illustrates the corresponding edit lattice of matrix \ref{levenshteinMatrix}. Each edge is weighted by the sum of the costs of its parts. Each unique part has a unit cost. The goal is to determine the sequence of edits in the lattice structure that has the maximum overlap with the gold-standard annotation. To accomplish this, we change the cost of those edges that have an equivalent match to the gold-standard to $-(u+1)\times |E|$, e.g. the edge between 6,6 and 7,8. Parameter $u$ is defined as a threshold to avoid a substantially large number of unchanged edits. Transitive edits are allowed in the range of this threshold which is set to 2 in this lattice. 

By performing a shortest path algorithm on the lattice, e.g. using Bellman-Ford algorithm, we obtain the shortest path in the lattice (a single-source problem with negative edges). Dahlmeier et al. \cite{dahlmeier2012better} proved that the edits with the maximum overlap with the gold-standard edits are those of the shortest-path. 

Back to our example, the $M^2$ Scorer evaluates the correction system with $Precision=0.5000$, $Recall=0.5000$ and $F_0.5=0.5000$ which seems to be justifiable since the correction system has predicted only one correction that exists also in the gold-standard corrections. 

Although $M^2$ Scorer is currently a standard metric in evaluating error correction systems, having  been  used  to  rank  error correction systems in the 2013 and 2014 CoNLL  \cite{kao2013conll,ng2014conll} and EMNLP 2014 \cite{mohit2014first} shared tasks, it also has some weak points. In experimenting with the baseline system, since there is theoretically no correction to be done, the results of $M^2$ Scorer appear to be non-interpretable ($P=100\%, R=0, F_{0.5}=0$). On the other hand, partial edits are also ignored. Although the system prediction $P=$\{more $\rightarrow$ the more\} includes a part of the gold-standard annotation, if assumed as the only predicted edit by the correction system, it is evaluated with no difference with a completely incorrect prediction ($P=0, R=0, F_{0.5}=0$). These limitations have been the motivation to present the standard evaluation method for error correction systems that we discuss in the following section.

\subsection{I-measure}
\label{imeasure}

I-measure is a novel method in response to the limitations of the previous methods, particularly the $M^2$ Scorer \cite{felice2015towards}. Unlike $M^2$ Scorer that uses phrase-level edits, I-measure uses tokens as the unit of evaluation. It also provides scores for both detection and correction tasks and it is sensitive to different types of edit corrections. I-measure designs a new evaluation system by presenting a new format for the gold-standard annotations. Then, by determining an optimal alignment  between an input, a prediction and a gold-standard annotation, it computes the metrics that evaluates the performance of the correction system.

\begin{lrbox}{\myXMLbox}
\begin{lstlisting}[style=listXML]
<?xml version='1.0' encoding='UTF-8'?>
<scripts>
	<script id="1">
		<sentence id="1" numann="1">
			<text>
				And we keep track of all family members health conditions .
			</text>
			<error-list>
				<error id="1" req="yes" type="Vm">
					<alt ann="0">
						<c end="2" start="2">can</c>
					</alt>
				</error>
				<error id="2" req="yes" type="Mec">
					<alt ann="0">
						<c end="8" start="8">'</c>
					</alt>
				</error>
				<error id="3" req="yes" type="Rloc-">
					<alt ann="0">
						<c end="10" start="9" />
					</alt>
				</error>
			</error-list>
		</sentence>
	</script>
</scripts>
\end{lstlisting}
\end{lrbox}

\newsavebox\myv

\begin{lrbox}{\myv}\begin{minipage}{\textwidth}
\begin{verbatim}
S And we keep track of all family members 
health conditions .
A 2 2|||Vm|||can|||REQUIRED|||-NONE-|||0
A 8 8|||Mec|||'|||REQUIRED|||-NONE-|||0
A 9 10|||Rloc-||||||REQUIRED|||-NONE-|||0
\end{verbatim}
\end{minipage}\end{lrbox}

\begin{figure}[!h]
\centering
\scalebox{.95}{
\begin{minipage}{\linewidth}
      \begin{minipage}{0.25\linewidth}
      \resizebox{2.7\textwidth}{!}{\usebox\myv}
      \end{minipage}
      \hspace{0.05\linewidth}
      \begin{minipage}{0.7\linewidth}
          \begin{center}
			\hspace*{3cm}\scalebox{0.7}{\usebox{\myXMLbox}}
			\end{center}
      \end{minipage}
  \end{minipage}
}
\label{fig_imeasure_annotation}
\caption{An example of conversion of the gold-standard \textit{m2} format to the annotation scheme of I-measure}
\end{figure}

\subsubsection{Method}

In the first step, I-measure defines a new scheme for the gold-standard annotations where each sentence is annotated with a set of errors and their possible corrections. Figure \ref{fig_imeasure_annotation} illustrates the equivalent scheme of a gold-standard annotation example from the CoNLL 2014 Shared Task corpus. Each error is detected based on the location. Since all the correction alternatives are mutually exclusive, different mixture of corrections from different annotators lead to a better estimation of the evaluation.

Using dynamic  programming  implementation  of  the  Sum  of  Pairs   alignment, a globally optimal alignment between the source sentence, the predicted output and the gold-standard annotation is created. Sum of Pairs method is a common method in Bioinformatics that scores a multiple sequence alignment by summing the scores of each pairwise alignment. These scores are based on of three constrains in error correction:

\begin{itemize}[noitemsep]
\item $c_{match}$: cost of matches
\item $c_{gap} > c_{match}$: cost of insertions and deletions
\item $c_{mismatch} > c_{gap}$: cost of substations
\item $2c_{gap} > c_{mismatch}$: to ensure that the desired alignment has a lower cost than a initial alignment
\end{itemize}

Once the optimal alignment found, the evaluation is calculated using an extended version of the  Writer-Annotator-System (WAS) introduced in \cite{chodorow2012problems} and is shown in Table \ref{was_method}. In this table, a, b and c are used to indicate the agreements, e.g., if the source, the hypothesis and the gold-standard annotation are token a, a is a true-negative detection and correction. The cases where $Source\neq Hypothesis \neq Gold-standard$ belong to false-positive and false negative classes. These cases are specific by FPN class.

\begin{table}[h]
\centering
\scalebox{.7}{
\begin{tabular}{|l|l|l||l|l|}
\hline
\multicolumn{3}{|l|}{Tokens}        & \multicolumn{2}{l|}{Classification} \\ \hline \hline
Source & Hypothesis & Gold-standard & Detection      & Correction         \\ \hline
a      & a          & a             & TN             & TN                 \\ \hline
a      & a          & b             & FN             & FN                 \\ \hline
a      & a          & -             & FN             & FN                 \\ \hline
a      & b          & a             & FP             & FP                 \\ \hline
a      & b          & b             & TP             & TP                 \\ \hline
a      & b          & c             & TP             & FP, FN, FPN        \\ \hline
a      & b          & -             & TP             & FP, FN, FPN      \\ \hline
a      & -          & a             & FP             & FP                 \\ \hline
a      & -          & b             & TP             & FP, FN, FPN        \\ \hline
a      & -          & -             & TP             & TP                 \\ \hline
-      & a          & a             & TP             & TP                 \\ \hline
-      & a          & b             & TP             & FP, FN, FPN        \\ \hline
-      & a          & -             & FP             & FP                 \\ \hline
-      & -          & a             & FN             & FN                 \\ \hline
\end{tabular}
}
\caption{Extended Writer-Annotator-System evaluation system}
\label{was_method}
\end{table}

Traditionally, NLP-related systems are evaluated using a 2$\times$2 contingency table that provides a comparison between system outputs and the gold-standard annotations. Considering the input text in the evaluation, the WAS table is an alternative to represent the $2\times2\times2$ possible combinations in correction system evaluation.  

Finally, evaluation metrics are calculated using $precision$, $recall$ and $F-measure$, defined in section \ref{sec_stan_metrics}. In order to reward correction more than preservation (i.e., "doing nothing"), a weighted version of accuracy is also defined:

\begin{equation}
WAcc=\frac{w.TP+TN}{w.(TF+FP) + TN + FN - (w+1).\frac{FPN}{2}}
\end{equation}

where $w>1$.

\subsection{BLEU and GLEU}
\label{bleugleuscores}

One of the most widely used automatic evaluation metrics is BLEU score \cite{papineni2002bleu}. It is computed as the geometric  mean  of  the  modified n-gram $precision$, multiplied by a brevity penalty, $\rho$, to control for recall by penalizing short translations:

\begin{equation}
BLEU = \rho(\prod_{i=1}^{N}precision_{i})^{\frac{1}{N}}
\end{equation}

where $n$ is the n-gram order that is most often 4, and $precision_i$ and $\rho$ are calculated as follow:

\noindent\begin{minipage}{.6\linewidth}
\begin{equation}
Precision_{i} = \frac{\sum_{t_i}min\{{C_h(t_i),max_jC_{hj}(t_i)\}}}{H(i)}
\end{equation}
\end{minipage}
\begin{minipage}{.4\linewidth}
\begin{equation}
\rho = exp\{min(0, \frac{n-L}{n})\}
\end{equation}
\end{minipage}
\vspace{2mm}

where $H(i)$ is the number of $i$-gram tuples in the hypothesis, $t_i$ is an $i$-gram tuple in hypothesis $h$, $C_h(t_i)$ is the number of times $t_i$ occurs in the hypothesis and $C_{hj}(t_i)$ is the number of times $t_i$ occurs in the gold-standard annotation $j$ of this hypothesis. In calculating brevity penalty $\rho$, $n$ refers to the length of the hypothesis and $L$ refers to the length of the gold-standard annotation.

Recently, Napoles et al. ameliorated BLEU metric for evaluation of grammatical error correction systems and proposed the \emph{Generalized Language Evaluation Understanding} (GLEU) \cite{napoles2015ground,napoles2016gleu}. Similar to the I-measure metric, the precision calculation is modified to assign extra weight to n-grams that are present in the gold-standard annotation and the system prediction, but not those of the input (set of n-grams R$\backslash$S), and it also penalizes the grams in the candidates that are present in the source but not the gold-standard annotation (S$\backslash$R). This metric is defined as:

\begin{equation}
{p_n}' = \frac{\sum_{n-gram \in C}count_{R\backslash S}(n-gram)-\lambda(count_{S\backslash R}(n-gram))+count_{R}(n-gram)}{\sum_{n-gram\in C}count_S(n-gram)+\sum_{n_gram\in R\backslash S}count_{R\backslash S}(n-gram)}
\end{equation}

where $\lambda$ refers to the rate of penalty for incorrectly changed $n_grams$, and function $count$ is calculated for a set of n-grams $B$ as:

\begin{equation}
count_B(n-gram)=\sum_{{n-gram}'\in B} d(n-gram,{n-gram}')
\end{equation}

where $d(n-gram,{n_gram}')$ has a unit value in the case that $n-gram$ and ${n-gram}'$ are equal. Using these parameters, GLEU(C,R,S) score for a candidate $C$ for input $S$ and reference $R$ is defined as:

\begin{gather}
BP = \left\{\begin{matrix}
1 & c>  r\\ 
e^{(1-c)/r} & c\leq r
\end{matrix}\right.\\
GLEU(C,R,S)=BP.exp(\sum_{n=1}^{N}w_nlog{p_n}')
\end{gather}

Since the task of evaluation in machine correction systems is limited to a comparative method between the prediction results and the gold-standard annotations, some of the evaluation measures in machine translation systems, e.g., manual evaluation for adequacy and fluency, does not appear to be pertinent in the case of machine correction systems. Results are reported in section \ref{Evaluation} using these metrics.

\chapter{Experimental Setup}
\label{Experimental_Setup}

This chapter discusses the experimental procedure used to create our error correction system. First, in section \ref{sec_task}, we present our task and the pipeline the we followed in developing the project. Section \ref{sec_qalb_corpus} introduces the QALB corpus used for our experiments. In this section, the corpus is preprocessed and different details are analyzed. Finally, section \ref{sec_technical_detials} contains the technical details for the implementation of the project. 

\section{Task}
\label{sec_task}

Conceptually, the task of creating our error correction system can be classified in three modules:

\begin{enumerate}
\item{\textbf{Modeling}}: Given a parallel corpus of potentially incorrect words versus their corrected forms, we define a model by defining the representation of data and the approach of the correction.
    
\item{\textbf{Training}}: In order to estimate parameters $\theta$ that most most suite to the corpus, we need to train a method on the training data. 

\item{\textbf{Searching}}: In this last step, we solve the selection mechanism $argmax$ by searching for the hypothesis with the highest probability.
\end{enumerate}

Although these three models are practically addressed in chapter \ref{Method}, this chapter covers a partial description regarding the QALB corpus that is used in the training and evaluating of the models.

\section{The QALB corpus}
\label{sec_qalb_corpus}

Qatar Arabic Language Bank (QALB)\footnote{\url{http://nlp.qatar.cmu.edu/qalb/}} is created as a part of a collaborative project between Columbia University and the Carnegie Mellon University Qatar, funded by the Qatar National Research Fund. In our research, we used the release of QALB at ANLP-ACL 2015 which includes data sets of native and non-native (L2 data sets) Arabic speakers. \cite{zaghouani2014large,obeid2013web}. 

QALB corpus is provided in three subsets: training, development and test, for which (\textit{*.train.*}), (\textit{*.dev.*}) and (\textit{*.test.*}) are respectively used as the file extension. This is not the case for the non-native corpus(\textit{L2}) where the training and development sets are only provided. To evaluate different models on the development data, QALB corpus comes with some evaluation scripts that may be useful to evaluate the performance of different models. However, we have not used them in this project.

Three versions of the data are included in the corpus: \textit{*.m2} is the format accepted by the scorer for evaluation including the gold annotations, \textit{*.sent} includes plain documents with identifiers, and finally \textit{*.column} provides the feature files in column format generated using \cite{pasha2014madamira}.

The QALB corpus is composed of machine translation texts, human user comments, weblog contents, student essays and non-native  essays. Their goal in the annotation is to correct all the errors by maintaining the minimum edit actions. Corrected errors in the QALB corpus can be classified in the following six classes:
\begin{enumerate}[noitemsep]
\item Spelling errors: common typographical errors
\item Punctuation errors
\item Lexical errors: inadequate lexicon
\item Morphological errors: incorrect derivation or inflection
\item Syntactic errors: grammatical errors, e.g., agreement of gender, number, definiteness
\item Dialectal errors: specific words are detected as incorrect if not present in the Almaany reference dictionary \footnote{\url{http://www.almaany.com/}}.
\end{enumerate}

\subsection{QALB corpus structure}
\label{processing_QALB}

Using the annotation style of the CoNLL-2013 shared task in \cite{kao2013conll}, the QALB corpus is consists of blocks of annotated phrases. In each block, the first line starting by '\textit{S}' is a document token that may encompass a single sentence or a paragraph of different sentences. Following the sentence line, the corrected form of each token appears in lines starting by '\textit{A}'. In each line containing a correction, three different pieces of information are important to create our data set:
\begin{enumerate}[noitemsep]
\item location of each token represented by $beginning-ending$ identifiers (indexed from 0). This location refers to the spaces between each token
\item correction type needed to correct a given token
\item the corrected form of the token
\end{enumerate} 
Each block is separated by an empty line.

\begin{figure}[h]
    \centering
    \includegraphics[scale=0.8]{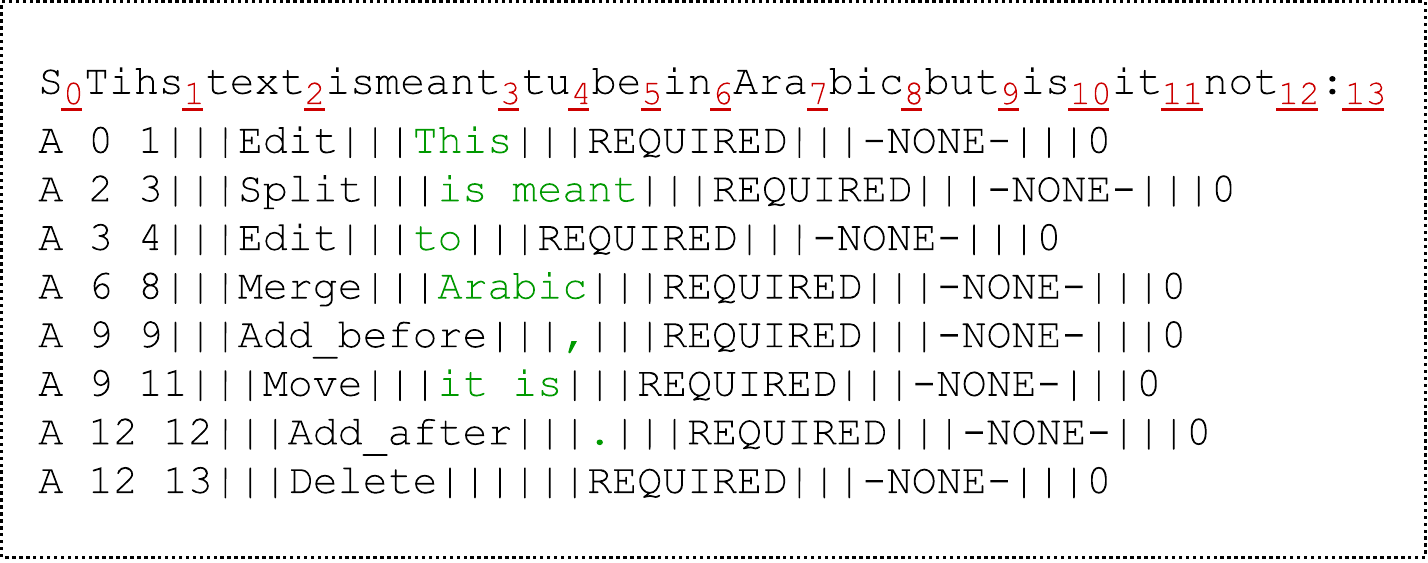}
    \caption{An example of the correction process based on the structure of the QALB corpus. $S$ and $A$ refer respectively to the beginning of the sentence and the gold annotators' correction}
   \label{fig:qalBexample}
\end{figure}

Figure \ref{fig:qalBexample} demonstrates a correction  annotation  based on the QALB corpus for a dummy writer's text. In this example, the phrase is started by $S$ and then the correction annotations are followed in $A$ lines. The first correction concerning the token that starts from location 0 to location 1, i.e. $Tihs$, is to be corrected to $This$ by an $Edit$ action. The token in location 2 to 3 needs a $Split$ action to be corrected as $is$ $meant$ instead of $ismeant$. Following all the $A$ lines, these annotations yield the correct phrase \textit{"This text is meant to be in Arabic but, it is not."}.  Locations are detectable based on the space characters in the phrases, which are enumerated in red in this example. The last index $A$ lines, i.e., $|||0$, indicate the number of annotators. This may influence the evaluation results depending on the metric. However, for the current version of the QALB corpus, we could not find more than one annotator. An example of the QALB corpus can be found in \ref{app_qabl_sample}. 

If we consider each action as a function $\forall x_i, x_{i+n} Correct(x_i, x_{i+n})$, we can define the correction actions in the QALB corpus as follows:

\begin{itemize}[noitemsep]
\item \textit{Add\textunderscore before}: insert a token in front of another token ($n=0$). 
\item \textit{Merge}: merge multiple tokens ($n \geq 2$). The maximum value that we observed for this action was 4.
\item \textit{Split}: split a token into multiple tokens ($n=1$).
\item \textit{Delete}: delete a token ($action=""$ and $n=1$). 
\item \textit{Edit}: replace a token with a different token ($n=1$).
\item \textit{Move}: move a token to a different location in the sentence ($n \geq 2$). The maximum value that we observed for this action was 7.
\item \textit{Add\textunderscore after}: insert a token after another token. This action is always accompanied by a $Delete$ action ($n=0$). 
\item \textit{Other}: a complex action that may involve multiple tokens ($n \geq 1$). The maximum value that we observed for this action was 6.
\end{itemize}

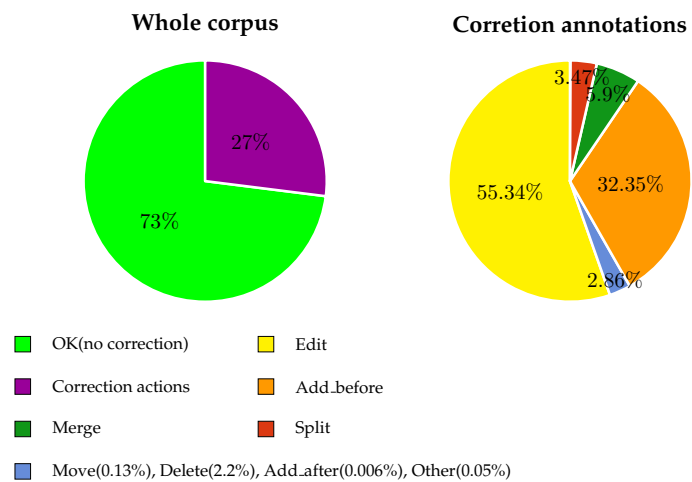
\begin{figure}[h]
    	\centering
    	\scalebox{.8}{
\begin{tikzpicture}
[
    pie chart,
    slice type={OK}{green},
    slice type={Edit}{yellow},
    slice type={split}{rosso},
    slice type={merge}{verde},
    slice type={other}{blu},
    slice type={ETC}{viola},
    slice type={Add_before}{giallo},
    slice type={One}{gray},
    slice type={moreThanOne}{orange},
    pie values/.style={font={\small}},
    scale=2
]

\pie[xshift=1.5cm,values of coltello/.style={pos=1.1}]{Whole corpus}{73/OK, 27/ETC} 
\pie[xshift=4.5cm,values of coltello/.style={pos=1.1}]{Corretion annotations}{55.34/Edit, 2.86/other, 32.35/Add_before, 5.9/merge, 3.47/split} 

\legend[shift={(0cm,-1cm)}]{{OK(no correction)}/OK, {Correction actions}/ETC, {Merge}/merge, {Move(0.13\%), Delete(2.2\%), Add$\textunderscore$after(0.006\%), Other(0.05\%)}/other}
\legend[shift={(2cm,-1cm)}]{{Edit}/Edit, {Add$\textunderscore$before}/Add_before, {Split}/split}





\end{tikzpicture}
}
\caption{Actions used by annotators for correction in the QALB corpus}
\label{fig:QALB_stat}
\end{figure}

Figure \ref{fig:QALB_stat} shows the pie charts of the correction actions proportionally. The 73\% of the tokens in the QALB corpus that are not annotated are considered correct, i.e., $action=OK$. The proportion of each action is as follows:  $Edit$: 169 769, $Move$: 427, $Add\textunderscore before$: 99258, $Merge$: 18267, $Delete$: 6778, $Split$: 10675, $Other$: 1563 and $Add\textunderscore after$: 20, which are illustrated in the pie chart on the right. 


\subsection{Creating data sets}

Based on the level of modeling, we need to manipulate the corpus in order to extract the needed data. For instance, we can create our data set based on three pieces of information: \textbf{original tokens} ($S$ line), \textbf{corrected tokens} and the \textbf{correction actions} (both in $A$ line). With reference to the example in figure \ref{fig:QALB_stat}, the structure of the data set of this block will be:

\begin{lstlisting}[caption=Processing the annotation block of figure \ref{fig:QALB_stat}]
[
	[["Tihs", "This", "Edit"], 
	["text", "text", "OK"], 
	["ismeant", "is meant", "Split"],
	["tu", "to", "Edit"],
	["in", "in", "OK"],
	["Ara bic", "Arabic", "Merge"],
	["but", "but", "OK"],
	["", ",", "Add_before"],
	["is it", "it is", "Move"],
	["", ".", "Add_after"],
	[":", "", "Delete"]]
]
\end{lstlisting}

One of the challenges was the case of having more than one action for a given token. This amounts to 701 tokens requiring 306 757 correction actions. Thus, for juxtaposing actions to the tokens, in the case of a $1 \rightarrow 1$ juxtaposition between an annotation and a token, each action would be assigned as the correction action of the token. However, if more than one correction action is available for a given token, we concatenate all the actions into one action. Since we have not found two different actions for a single token in the current version of QALB, we simply concatenate the corrections keeping the same action.

\begin{figure}[h]
    \centering
    \includegraphics[scale=1]{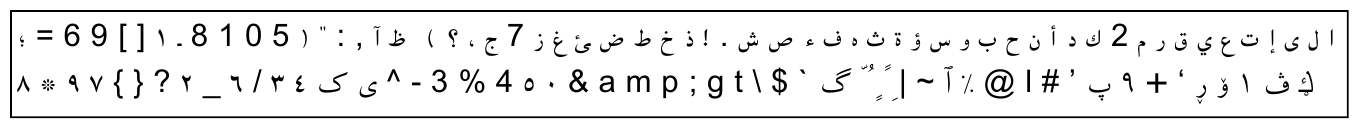}
    \caption{Existing characters in the QALB corpus, each one separated by a space}
    \label{fig:characters}
\end{figure}

After analyzing different aspects of the corpus, we created different data sets to train our models. The corpus has a density of 15.80 errors/block and it has 115 characters (including space) from all 1 022 126 tokens in the corpus. We have also considered $<EOS>$ as a specific character with which to end each string. The role of this character is explained in the next chapter. Figure \ref{fig:characters} indicates all the 115 extracted characters of the QALB corpus. It seems that $\&gt;$ and $\&amp;$ are not pre-processed to be converted to \& and $>$ respectively.

One of the important points that will emerge further during the evaluation process, is when the system yields a correction phrase with a different size from that of the source. The remarkable difference between 98577 empty strings among the original tokens and 6778 empty tokens among the corrected tokens, indicates the variable size of the source/target texts.

\subsection{Refining data sets}

Although the curse of dimensionality does not hit the character-level models in general, due to the limited number of existing characters in a corpus, it can create serious problems for the word-level models, both in performance and processing. 

We have refined the data sets to obtain a more optimized model. In this technique, we refine the parameters of the model based on the frequency of the data. This technique is more usable when the model is trained in a word-level in which multiple matrices in large dimensions e.g., $10^6 \times 10^2$ are presented.

\begin{figure}[!h]
    	\centering
\begin{tikzpicture}[scale=0.7]
    \begin{axis}[
        width  = \textwidth,
        axis y line*=left,
        axis x line=bottom,
        height = 9cm,
        major x tick style = transparent,
        ybar=5*\pgflinewidth,
        bar width=20pt,
        ymajorgrids = true,
        ylabel = {Number of words},
        xlabel = {Frequency},
        symbolic x coords={1,2,3,4,5,6,7,8,9-14,15+},
        xtick = data,
            scaled y ticks = false,
            enlarge x limits=0.1,
            axis line style={-},
            ymin=0,ymax=50998,
    ]
        \addplot[style={blue,fill=blue,mark=none}]
        	coordinates {(1, 50998) (2, 43743) (3, 6631) (4, 11186) (5, 3082) (6, 5284) (7, 1814) (8, 3068) (9-14, 7553) (15+, 13251)};

    \end{axis}
\end{tikzpicture}
\caption{Number of words based on frequency in the QALB corpus}
\label{fig:freq_num}
\end{figure}
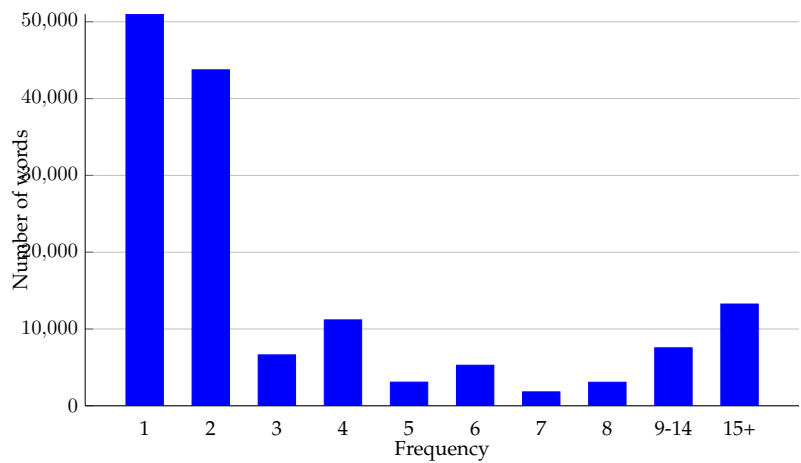

Among the entire tokens in the corpus, 146610 are unique tokens, which equate to 14.18\%. A total of 789 different frequencies in the non-sequential range of [1, 99808] exist. The frequency of the words with 1 to 14 occurrences in the corpus comprises 91\% of the words. Thus, the remaining tokens have more than 14 occurrences. Figure \ref{fig:freq_num} summarizes the frequency of the words in the QALB corpus.

\section{Technical details}
\label{sec_technical_detials}

In this project, we implemented our models in the Python version of DyNet \footnote{It is released open-source and available at \url{https://github.com/clab/dynet}.}. The Dynamic Neural Network Toolkit, or DyNet \cite{neubig2017dynet}, is a neural network library suited to networks that have dynamic structures. DyNet supports both static and dynamic declaration strategies used in neural networks computations. In the dynamic declaration, each network is built by using a directed and acyclic computation graph that is composed of \texttt{expressions} and \texttt{parameters} that define the model. 

Working efficiently on CPU or GPU, DyNet has powered a number of NLP research papers and projects recently. 
\chapter{Correction models}
\label{Method}

This chapter will describe how experiments were implemented within this research and what parameters were used in order to carry them out. We also present the training details of our models for the task of error correction. Results of these models are evaluated in chapter \ref{Evaluation}.

\section{Sentence Representation}

The first step in creating a model, is to choose a representation for the input and output. For instance, a sentence can be represented in word-level, which is transferring the data with indices that refer to the words. However, in this work, we anticipate the need to handle an exponentially large input  space  of tokens by  choosing  a character-level sentence representation. Thus, model inputs are individual characters which are a sequence of indices corresponding characters. We have defined a specific character $<EOS>$ in order to specify the end of each sentence. This character will be later used in the training process as a criteria. Note that in the character-level representation, we have also considered space as a unique character, i.e. 116 unique characters in all \ref{fig:characters}.

Once a sentence is mapped into indices, the characters are embedded \cite{bengio2003neural}, i.e. each character is represented by a one-hot vector and is multiplied by a trainable matrix with the \texttt{size of the input} $\times$ \texttt{embeddings size}. The embedding allows the model to group together characters that are similar for the task. These embeddings are fed into the models.

\section{Models}
Previously in chapter \ref{Background}, we reviewed a background of the models that we're using in this research. In this section, we present the technical points of the models that we have used. Having said that we explored different hyper-parameters, here we present those for which the experiments are done. 

\subsection{RNN model}

\begin{figure}[h]
\centering
\begin{tikzpicture}[
  hid/.style 2 args={rectangle split, rectangle split horizontal, draw=#2, rectangle split parts=#1, fill=#2!20, outer sep=1mm},
  label/.style={font=\small}]
    \tikzset{>=stealth',every on chain/.append style={join},
      every join/.style={->}}  

  \foreach \step in {1,2,3}
    \node[hid={4}{cyan}] (h\step) at (3*\step, 0) {};

  \foreach \step in {1,2,3}
    \node[hid={4}{gray}] (w\step) at (3*\step, -1.5) {};    
    
  \foreach \step in {1,2,3}
    \draw[->] (w\step.north) -> (h\step.south);

  \foreach \step/\next in {1/2,2/3}
    \draw[->] (h\step.east) -> (h\next.west);

   \node[] (h0) at (0.6, 0) {};
   \node[] (h4) at (11.2, 0) {};
    \draw[->] (h0.east) -> (h1.west);
    \draw[->] (h3.east) -> (h4.west);
 
  \node[hid={1}{orange}] (o1) at (3, 1.5) {$softmax$};
  \node[hid={1}{orange}] (o2) at (6, 1.5) {$softmax$};
  \node[hid={1}{orange}] (o3) at (9, 1.5) {$softmax$};
  
  \draw[->] (h1.north) -> (o1.south);
  \draw[->] (h2.north) -> (o2.south);
  \draw[->] (h3.north) -> (o3.south);
  
  \draw[->] (o1.north) -> (3,2.5);
  \draw[->] (o2.north) -> (6,2.5);
  \draw[->] (o3.north) -> (9,2.5);
  
  \node[label]  at (3, 3){$o_1$};
  \node[label]  at (6, 3){$o_2$};
  \node[label]  at (9, 3){$o_3$};
  
  \draw[->] (3,-2.5) -> (w1.south);
  \draw[->] (6,-2.5) -> (w2.south);
  \draw[->] (9,-2.5) -> (w3.south);
  
  \node[label]  at (3, -3){$x_1$};
  \node[label]  at (6, -3){$x_2$};
  \node[label]  at (9, -3){$x_3$};

  \foreach \step in {1,2}
    \node[label] (hlabel\step) at (3*\step+0.5, -0.4) {$\mathbf{h}_\step$};    
    
    \node[label] (hlabel3) at (10.6, -0.4){$\mathbf{h}_3$};
    

  \foreach \step in {1,2,3}
    \node[label] (wlabel\step) at (3*\step+1.2, -2) {$embedding\step$};
        
\end{tikzpicture}
\caption{A recurrent neural network for error correction. The input is a potentially erroneous sequence. The output of the model is supposed to be the corrected form of the input. Note the fix size of the input and output sequence.}
\label{fig_rnn_correction}
\end{figure}
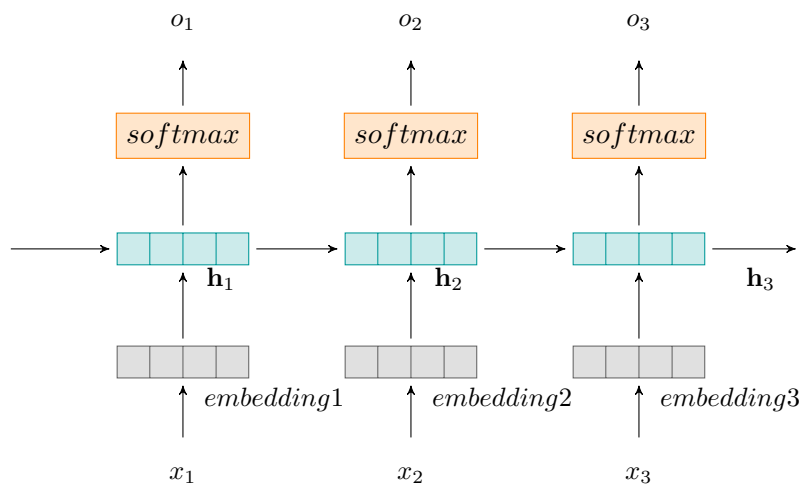

The first model that we examined is an RNN as the simplest architecture. By passing the embedding of the input sentence as the input of the network, the corrected form of the input is provided as the output of each state of the network. We use softmax layer to produce probability distribution over output
alphabet at each time step. Figure \ref{fig_rnn_correction} illustrates this process. 

The hyper-parameters of the network are:

\begin{itemize}[noitemsep]
\item Architecture: LSTM
\item Number of layers: 2
\item Cells in each layer: 80
\item Number of embeddings: 50
\item Parameters initialization: [-0.1, 0.1]
\item Number of epochs: 20
\item Learning rate: Default value of \texttt{SimpleSGDTrainer} in DyNet, i.e., $\eta_t = \frac{\eta_0}{1+\eta_{decay}t}$ at epoch $t$.
\end{itemize}

We train the model by minimizing the negative log likelihood of the training data using stochastic gradient. The codes of the error calculation fo each model is available in the appendix \ref{error_calculation}.

\subsection{BRNN model}

The structure of our BRNN model is exactly the same with the RNN model. The only difference is that an bidirectional RNN consists of two unidirectional RNNs, one reading the inputs in forward direction and the other in backward direction. So, a sum of forward and backward RNN outputs is the output of bidirectional RNN. The hyper-parameters of our model is the following:

\begin{itemize}[noitemsep]
\item Architecture: LSTM
\item Number of layers: 4
\item Cells in each layer: 90
\item Number of embeddings: 50
\item Parameters initialization: [-0.1, 0.1]
\item Number of epochs: 20
\item Learning rate: Default value of \texttt{SimpleSGDTrainer} in DyNet, i.e., $\eta_t = \frac{\eta_0}{1+\eta_{decay}t}$ at epoch $t$.
\end{itemize}

BRNN in addition to the previous observations, can also take next observations into account. This enables it to be more efficient in dealing with all kinds of errors.

\subsection{Encoder-decoder model}

Using two RNN models, we created an encoder-decoder model with the following hyper-parameters:

\begin{itemize}[noitemsep]
\item Architecture: LSTM
\item Number of encoder layers: 4
\item Number of decoder layers: 4
\item Cells in each layer of the encoder: 100
\item Cells in each layer of the decoder: 100
\item Number of embeddings: 50
\item Parameters initialization: [-0.1, 0.1]
\item Number of epochs: 10
\item Learning rate: Default value of \texttt{SimpleSGDTrainer} in DyNet, i.e., $\eta_t = \frac{\eta_0}{1+\eta_{decay}t}$ at epoch $t$.
\end{itemize}

\subsection{Attention-based encoder-decoder}

We added an attention mechanism between the encoder and the decoder model. This mechanism helps the decoder to pick only the encoded inputs that are important for each step of the decoding process. Once we calculate the importance of each encoded vector, we normalize the vectors using softmax and multiply each encoded vector by its weight to obtain a "time dependent" input encoding which is fed to each step of the decoder RNN. The only difference of this model with the encoder-decoder model in terms of hyper-parameters, is the number of epochs. We trained the model in 7 epochs.

\section{Output generation}
For an error correction system, sentences with spelling or grammatical errors should have lower probability than
their corrected versions, since they are less present in the whole corpus. Based on this assumption, as we explained in the description of the models, an output is the most probable element in the softmax probability distribution (demonstrated in algorithm \ref{algo_char2char_rrn}). Although, this is not the only approach to generate an output. Dahlmeier and Ng \cite{dahlmeier2012beam} developed a \emph{beam-search} decoder to iteratively generate sentence-level candidates and ranking them.

\begin{algorithm}
\caption{Correction of an input sequence using a character-level model(RNN or BRNN)}
\label{algo_char2char_rrn}
\hspace*{1cm} \textbf{Input}: Input sequence $C\gets c_1,c_2,...,c_N$\\
\hspace*{1cm} \textbf{Output}: Corrected sequence $O \gets o_1,o_2,...,o_N$
\begin{algorithmic}[1]

\Procedure{GenerateOutput}{$C$}
\State $embeddedInput \gets \texttt{embedSequence(C)}$
\State $modelOutputs \gets \texttt{runModel}(embeddedInput)$
\State $p_0,p_1,...,p_N \gets \texttt{getProbabilities}(modelOutputs)$

\State $index \gets \texttt{0}$
\While {$i < N$}
	\State $predictedIndex \gets \texttt{argmax}(p_i)$
	\State $o_i \gets \texttt{indexToCharacter}(predictedIndex)$
	\State $i \gets i+1$
\EndWhile

\Return $O$
\EndProcedure
\end{algorithmic}
\end{algorithm}

In the case of encoder-decoder and attention-based encoder-decoder, since the output size can be variant, we have used two different conditions. Algorithm \ref{algo_char2char_encoder} shows the generation algorithm in an encoder-decoder model or an attention-based encoder-decoder model:

\begin{algorithm}
\caption{Correction of an input sequence using a character-level model(encoder-decoder or attention-based encoder-decoder)}
\label{algo_char2char_encoder}
\hspace*{1cm} \textbf{Input}: Input sequence $C\gets c_1,c_2,...,c_N$\\
\hspace*{1cm} \textbf{Output}: Corrected sequence $O \gets o_1,o_2,...,o_M$
\begin{algorithmic}[1]

\Procedure{GenerateOutput}{$C$}
\State $embeddedInput \gets \texttt{embedSequence(C)}$
\State $modelOutputs \gets \texttt{runModel}(embeddedInput)$
\State $p_0,p_1,...,p_N \gets \texttt{getProbabilities}(modelOutputs)$

\State $index \gets \texttt{0}$
\State $MaxSize \gets N \times 2$
\While {$i < MaxSize$}
	\State $predictedIndex \gets \texttt{argmax}(p_i)$
	\State $o_i \gets \texttt{indexToCharacter}(predictedIndex)$
	\State $i \gets i+1$
	
	\If {$o_i="<\text{EOS}>"$} 
	\State $\textbf{break}$
	\EndIf
\EndWhile

\Return $O$
\EndProcedure
\end{algorithmic}
\end{algorithm}


\chapter{Experiments and Results}
\label{Evaluation}

In the previous chapter, we presented four models for the task of error correction. The details of the models were explained. In this chapter, the results obtained from the models are discussed. First we analyze the data set and the error rate of the gold-standard corpus. The evaluation metrics are then applied on the results of the models. As we explained in section \ref{Evaluation}, using different metrics can reflect a better interpretation of the evaluation of the results. 

\section{Data}

A sequence in input or output of a error correction system can be represented in different levels. In character-level, a sequence is processed character by character. When searching for errors, humans often consider a bigger sequence of characters at word-level. Clause-level, phrase-level, sentence-level and text-level are other common representations for modeling sequences. 

In our research, we worked at character-level where each character in an input sequence is mapped to an real-valued number and then it is embedded. In order to model linguistic dependencies in each sequence, we took every character into account, including space. This enables us to deal with different kinds of errors and a larger range of characters in each sequence. On the other hand, the output of the models are also at character level. We will later see that the fact that we train the methods using the whole sentence size reduces the accuracy of  the models remarkably.

\begin{table}[h]
\centering
\label{training_set_details}
\begin{tabular}{l|llll}
           & \#sentences & \#words & \#characters & Corrected tokens \\ \hline
train set      & 19411       & 1041537 & 10231305     & 29.45\%          \\
validation set & 1017        & 54880   & 537182       & 30.35\%          \\
test set       & 968         & 52290   & 510582       & 31.32\%         
\end{tabular}
\caption{Train, validation and test set of the QALB corpus.}
\label{table6_1}
\end{table}

Three data sets are provided in the QALB corpus: 
\begin{itemize}
\item Train set is used to build up our prediction model. Our algorithm tries to tune itself to find optimal parameters with the back-propagation. This set is usually used to create multiple algorithms in order to compare their performances during the validation phase. 
\item Validation set is used to compare the performances of the prediction parameters created based on the training set. We select the parameters that has the best performance.
\item Test set is a set of examples used only to assess the performance of a fully-trained classifier. In the RNN case, we would use the test set to estimate the error rate after we have chosen the final model. 
\end{itemize}
Table \ref{table6_1} summarizes each data set in details. We used these data sets with the given sizes.

\section{Baseline system}

We define the pair of source sentences and the gold-standard annotations as the baseline of the models. In this baseline, we assume that non of our implemented models intervene in the task of correction and only references are considered as correction. Simply saying, Baseline system is the system that makes no corrections on the input text. The baseline enables us to interpret the performance of each model in comparison to the default results. 

\section{Results of the models}

This section presents evaluation results of the models using the metrics that we introduced in section \ref{evaluation_metrics}. Since each of these metrics reflect a different aspect of evaluation of each model, comparing them together enables us to obtain a better description of the performance of the models.

\subsection{MaxMatch $M^2$}

As explained in detail in section \ref{m2_method}, the MaxMatch $M_ 2$ metric computes the sequence of phrase-level edits between the source text and the predicted text in a way that it maximizes  their overlap. The edits are then scored using precision $P$, recall $R$ and F-score $F_{0.5}$.

\begin{table}[h]
\centering
\scalebox{1}{
\begin{tabular}{|l|l|l|l|}
\hline
\multirow{2}{*}{Model} & \multicolumn{3}{l|}{$M^2$ scorer} \\ \cline{2-4} 
                       & P        & R        & $F_{0.5}$  \\ \hline
Baseline               & 1.0000 & 0.0000 & 0.0000    \\ \hline
RNN                    & 0.5397  & 0.2487  & 0.4373      \\ \hline
BiRNN                  & 0.5544  & 0.2943  & 0.4711      \\ \hline
Encoder-decoder        & \textbf{0.5835}  & \textbf{0.3249}  & \textbf{0.5034}      \\ \hline
Attention              & 0.5132  & 0.2132  & 0.4155      \\ \hline

\end{tabular}
}
\caption{Evaluation results of the models using MaxMatch $M^2$ metric. Bold numbers indicate the best method per measure.}
\label{maxmeasure_table}
\end{table}

Table \ref{maxmeasure_table} demonstrates the results of the models using MaxMatch $M^2$ metric. In the baseline system, the metric detects complete precision of the system. This is because in the baseline system the gold-standard annotations are directly applied as the hypothesis of a model. So, we can justify the zero value of the recall and the $F_{0.5}$ consequently. However, these scores are not as high as the baseline system for other models.

We set the value of the transitive edits in the range of 2 for calculating the corresponding lattice of each edit. We set also the value of $\beta=2$ in calculating the F-score.

In addition to the \texttt{m2} format of the gold-standard annotations which is specifically used for MaxMatch $M^2$ metric, the reference correction of the QALB corpus is provided also in raw text with \texttt{.sent} format. Thus, we could evaluate the models without taking any possible preprocessing noises into account. For instance, evaluating the baseline model, we could have the following results $P=0.0179, R=0.0004, F_{0.5}=0.0017$. Even if these values are not remarkably different from those of the reference correction, it indicate probable mismatching during the preprocessing of the \texttt{.m2} files of the corpus.

We have used the original implementation of the MaxMatch $M^2$ introduced in \cite{dahlmeier2012better}. \footnote{Available at \url{http://nlp.comp.nus.edu.sg/software/}}.

\subsection{I-measure}

I-measure metric is a method in response to the limitations of the MaxMatch $M^2$ metric. As we explained in section \ref{imeasure}, the I-measure determines an optimal alignment between an input, a hypothesis and a gold-standard text. In addition to the standard metrics introduced in section \ref{Evaluation}, in this metric an  Improvement (I) score  is  computed by  comparing  system performance with that of the baseline which leaves the original text uncorrected. Table \ref{imeasure_results} demonstrates the evaluation results of the trained models using I-measure metric.

\begin{table}[h]
\centering
\scalebox{0.7}{
\begin{tabular}{l|l|l|l|l|l|l|l|l|l|l|}
\cline{2-11}
\multirow{2}{*}{}                & \multicolumn{2}{l|}{Baseline} & RNN       &            & \multicolumn{2}{l|}{BRNN} & \multicolumn{2}{l|}{Encoder-decoder} & \multicolumn{2}{l|}{Attention} \\ \cline{2-11} 
                                 & Detection     & Correction    & Detection & Correction & Detection   & Correction  & Detection        & Correction        & Detection     & Correction     \\ \hline \hline
\multicolumn{1}{|l|}{TP}  & 0 & 0 &  11286 & 208 & 11800 & 223 & 12213 & 247 & 11557 & 201  \\ \hline
\multicolumn{1}{|l|}{TN}  & 39589 & 39589 & 20881 & 20881 & 20378 & 20378 & 22204 & 22204 & 21070 & 21070 \\ \hline
\multicolumn{1}{|l|}{FP} & 0 & 0 &  18730 & 29808 & 19220 & 30797 & 17410 & 29376 & 18533 & 29889 \\ \hline
\multicolumn{1}{|l|}{FN} & 98318 & 98318 & 87031 & 98109 & 86518 & 98095 & 86111 & 98077 & 86758 & 98114 \\ \hline
\multicolumn{1}{|l|}{FPN} & 0 & 0 &  0 & 11078 & 0 & 11577 & 0 & 11966 & 0 & 11356 \\ \hline
\multicolumn{1}{|l|}{P} & 100.00 & 100.00 &  37.60 & 0.69 & 38.04 & 0.72 & \textbf{41.23} & \textbf{0.83 } & 38.41 & 0.67 \\ \hline
\multicolumn{1}{|l|}{R} & 0.00 & 0.00 &  11.48 & 0.21 & 12.00 & 0.23 & 12.42 & 0.25 & 11.76 & 0.20 \\ \hline
\multicolumn{1}{|l|}{$F_{0.5}$} & 0.00 & 0.00 &  25.84 & 0.48 & 26.53 & 0.50 & 28.16 & \textbf{0.57 }& 26.43 & 0.46 \\ \hline
\multicolumn{1}{|l|}{Acc} & 28.71 & 28.71 &  23.32 & 15.29 & 23.33 & 14.94 & 24.95 & 16.28 & 23.66 & 15.42 \\ \hline
\multicolumn{1}{|l|}{$Acc_b$} & 28.71 & 28.71 &  28.71 & 28.71 & 28.71 & 28.71 & 28.71 & 28.71 & 28.71 & 28.71 \\ \hline
\multicolumn{1}{|l|}{WAcc} & 28.71 & 28.71 &  25.87 & 13.11 & 26.03 & 12.76 & 27.83 & \textbf{14.05} & 26.30 & 13.23 \\ \hline
\multicolumn{1}{|l|}{$WAcc_b$} & 28.71 & 28.71 &  28.71 & 28.71 & 28.71 & 28.71 & 28.71 & 28.71 & 28.71 & 28.71 \\ \hline
\multicolumn{1}{|l|}{I} & 0.00 & 0.00 & -9.87 & -54.32 & -9.32 & -55.54 & -3.06 & \textbf{-51.07} & -8.39 & -53.92 \\ \hline
\end{tabular}
}
\caption{Evaluation of the models using I-measure metric}
\label{imeasure_results}
\end{table}

The I-measure evaluates a prediction in terms detection and correction. However, we are interested only in the correction results, the detection results can also be informative about the quality of correction of a model. Similar to the evaluation results of the MaxMatch $M^2$ metric in table \ref{maxmeasure_table}, the baseline systems is completely precise in terms of correction, i.e., $F_{0.5}=0.00, P=100.00$. On the other hand, other models could not obtain competitive results. This is mainly because of the poor alignment between the source text, the hypothesis and the reference. In comparison to other models, the encoder-decoder demonstrates better performance with an Improvement score of $I=-51.07$ and a weighted accuracy of $WAcc=14.05$. The weighted version of accuracy $WAcc$ rewards correction more than preservation. 

We set the values of $WAcc$ and F-$\beta$ for this metric to 2 and 0.5 respectively. We used the original implementation of the I-measure introduced in \cite{felice2015towards}. \footnote{Available at \url{https://github.com/mfelice/imeasure/}}.

\subsection{BLEU and GLEU}

BLEU was one of the first automatic metrics used in measuring translation accuracy and has became one of most common metrics in machine translation systems evaluation \cite{papineni2002bleu}. In addition to this metric, we also use GLEU which is a simple variant of BLEU showing a better correlation with human judgments in the evaluation task \cite{napoles2015ground}. The following table shows the evaluation results of our correction models using BLEU and GLEU metrics.

\begin{table}[h]
\centering
\begin{tabular}{|l|l|l|}
\hline
\multirow{2}{*}{Model} & \multirow{2}{*}{BLEU score} & \multirow{2}{*}{GLEU score} \\
                       &                             &                             \\ \hline
Baseline    & 1.000000 &  1.000000\\ \hline
RNN         & 0.349212  & 0.313035                           \\ \hline
BiRNN        & 0.305852   &   0.269368                          \\ \hline
Encoder-decoder    & 0.327858   &   0.292956 \\ \hline
Attention       & 0.312947  &       0.287664                      \\ \hline
\end{tabular}
\caption{Evaluation of the models using BLEU and GLEU metrics}
\label{bleu_gleu_resutls}
\end{table}

As we expect, since the baseline system contains the gold-standard correction, the BLEU and the GLEU scores for the baseline system have the maximum value $1.00$. Using metric, the RNN model shows higher scores in comparison to other models, i.e., $BLEU=0.3492$ and $GLEU=0.3130$. Note that in the GLEU metric, the precision is modified to assign extra weight to the n-grams that are present in the reference and the hypothesis, but not those of the input. 

We have used the last update of the original implementation of the GLEU \cite{napoles2015ground} introduced in \cite{napoles2016gleu}. \footnote{Available at \url{https://github.com/cnap/gec-ranking}}.






%

\chapter{Conclusion and future work}
\label{Conclusion}

Thus far, the results of each method are presented in depth. In this last chapter, these results will be discussed in more detail based on the research aims. Furthermore, several ideas about future works in this domain are proposed.

\section{Limitations}

A comparative example of a source text, its gold-standard correction and output of trained model is illustrated in appendix \ref{app_qabl_sample}. We tagged manually four kinds of errors in different colors: the green tags refer to the incorrect tokens in the input which are predicted correctly by a trained model, the yellow tags are the incorrect input words, the orange tags are the incorrectly predicted tokens of an incorrect input word and finally the cyan tags show the incorrect prediction of a correct input token. 

The purpose of using these tags is to demonstrate different kinds of correction and the quality of correction by each model for the same symbol. For instance, the incorrect form of "\includegraphics[scale = 0.9, trim =0mm 2mm 0mm 0mm]{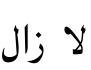}" is corrected by none of the models as "\includegraphics[scale = 0.9, trim =0mm 2mm 0mm 0mm]{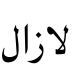}", while "\includegraphics[scale = 0.9, trim =0mm 2mm 0mm 0mm]{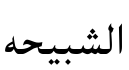}" is predicted correctly by the RNN, BRNN and encoder-decoder models as "\includegraphics[scale = 0.9, trim =0mm 2mm 0mm 0mm]{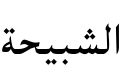}". Looking carefully at the distribution of incorrect prediction of correct input words (colored in cyan), one can deduce that the models perform less sensibly when the size the sequence become gradually bigger. To prove our observation, we tried to evaluate the models by limiting the sequences to a fixed size. 

\begin{table}[h]
\centering
\scalebox{0.8}{
\begin{tabular}{l|l|l|l|l|l|l|l|l|}
\cline{2-9}
\multirow{2}{*}{}                     & \multicolumn{2}{l|}{30}   & \multicolumn{2}{l|}{50}   & \multicolumn{2}{l|}{70}   & \multicolumn{2}{l|}{100}  \\ \cline{2-9} 
                                      & BLEU           & GLEU     & BLEU           & GLEU     & BLEU           & GLEU     & BLEU           & GLEU     \\ \hline \hline
\multicolumn{1}{|l|}{Baseline}        & 1.0000         & 1.0000   & 1.0000         & 1.0000   & 1.0000         & 1.0000   & 1.0000         & 1.0000   \\ \hline
\multicolumn{1}{|l|}{RNN}             & 0.575715 & 0.476466 & \textbf{0.579784 }& 0.545291 & 0.556254 & 0.543040 & 0.519782 & 0.514380 \\ \hline
\multicolumn{1}{|l|}{BRNN}            & 0.481971 & 0.397735 & 0.501957 & 0.469293 & 0.49238  & 0.480605 & 0.461176 & 0.455698 \\ \hline
\multicolumn{1}{|l|}{Encoder-decoder} & 0.471226 & 0.421491 & 0.488849 & 0.457512 & 0.479811 & 0.468159 & 0.456483 & 0.451533 \\ \hline
\multicolumn{1}{|l|}{Attention}       & 0.506216 & 0.389947 & 0.507188 & 0.477017 & 0.490448 & 0.478636 & 0.457459 & 0.452876 \\ \hline
\end{tabular}
}
\caption{Evaluation of models using fixed-size sequences}
\label{fixed_size_eval}
\end{table}

Table \ref{fixed_size_eval} demonstrates the BLEU and GLEU scores for a limited size of sequences of 30, 50, 70 and 100 characters. In comparison to the results of the models in table \ref{bleu_gleu_resutls}, this limitation shows considerably higher scores. 

We could not do the same experiment using MaxMatch nor I-measure methods. MaxMatch $M^2$ and I-measure need the gold-standard correction annotations for the evaluation, while BLEU and GLEU evaluate the models using the gold-standard raw text, without any specific information about the annotation.

\section{Future studies}

For the future studies, we have the following suggestions:

\begin{itemize}
\item Models to be explored in more levels, e.g., action-level, word-level and sentence-level. 
\item Limiting the length of the sequences in training models.
\item Using deeper networks with larger embedding size.
\item Preventing over-learning of models by not training them over correct input tokens (action ="OK").
\end{itemize}

\section{Conclusion}

The basic problem that the attention mechanism solves is that instead of forcing the network to encode all parameters into one fixed-length vector, it allows the network to take use of the input sequence. In other words, the attention mechanism gives the network access to the hidden state of the encoder which acts like an internal memory. Thus, the network is able to retrieve from the memory the related parameters for the current input. 

There is a tight relation between the attention mechanism and the memory mechanisms. For instance, the hidden state of a simple RNN by itself is a type of internal memory. However, the RNNs suffer from the vanishing gradient problem that hinders modeling long distance dependencies, which is a common phenomenon in human language. On the other hand, LSTM uses a gating mechanism that provides an explicit memory for deletions and updates.

Recently more researchers tend towards more complex memory structures. For instance, End-to-End Memory Networks \cite{DBLP:journals/corr/SukhbaatarSWF15} allow the network to read multiple times an input sequence in order to make an output and also to update the memory contents at each step. These models are used in question answering \cite{weston2015towards} and to language modeling successfully, since it is able to make multiple computational steps, also known as hop, over an input story.

Future researches may reveal a clearer distinction between attention and memory mechanisms. In the recent studies, reinforcement learning techniques have been also used for error correction task \cite{DBLP:journals/corr/SakaguchiPD17,DBLP:journals/corr/ZarembaS15}.

\appendix
\chapter{Appendix}
\label{appendix}

\section{Adam}
\label{adam}
Adaptive Moment Estimation (Adam) \cite{kingma2014adam} is a stochastic optimization method that computes learning rates for each parameter. In addition to storing an exponentially decaying average of past squared gradients $v_t$, Adam also maintains an exponentially decaying average of past gradients $m_t$. The step taken at each update is proportional to the ratio of these two moments for which $\beta_1$ and $\beta_2$ are respectively used as the exponential decay rates. We can define each parameter as follow: 

\begin{gather}
g_t=\triangledown H_t(\theta_t-1)\\
m_t=\frac{\beta_1m_{t-1}+(1-\beta_1)g_t}{1-\beta_1^2}\\
v_t =\frac{\beta_2v_{t-1}+(1-\beta_2)g_t^2}{1-\beta_2^2}\\
\theta_t = \theta_{t_1} - \alpha.\frac{m_t}{\sqrt{v_t}+\epsilon}
\end{gather}

where $\epsilon$ is a very small number to prevent division by zero.

\newpage

\section{Annotation example from the QALB corpus}
\begin{figure}[!h]
\centering
\includegraphics[scale=1]{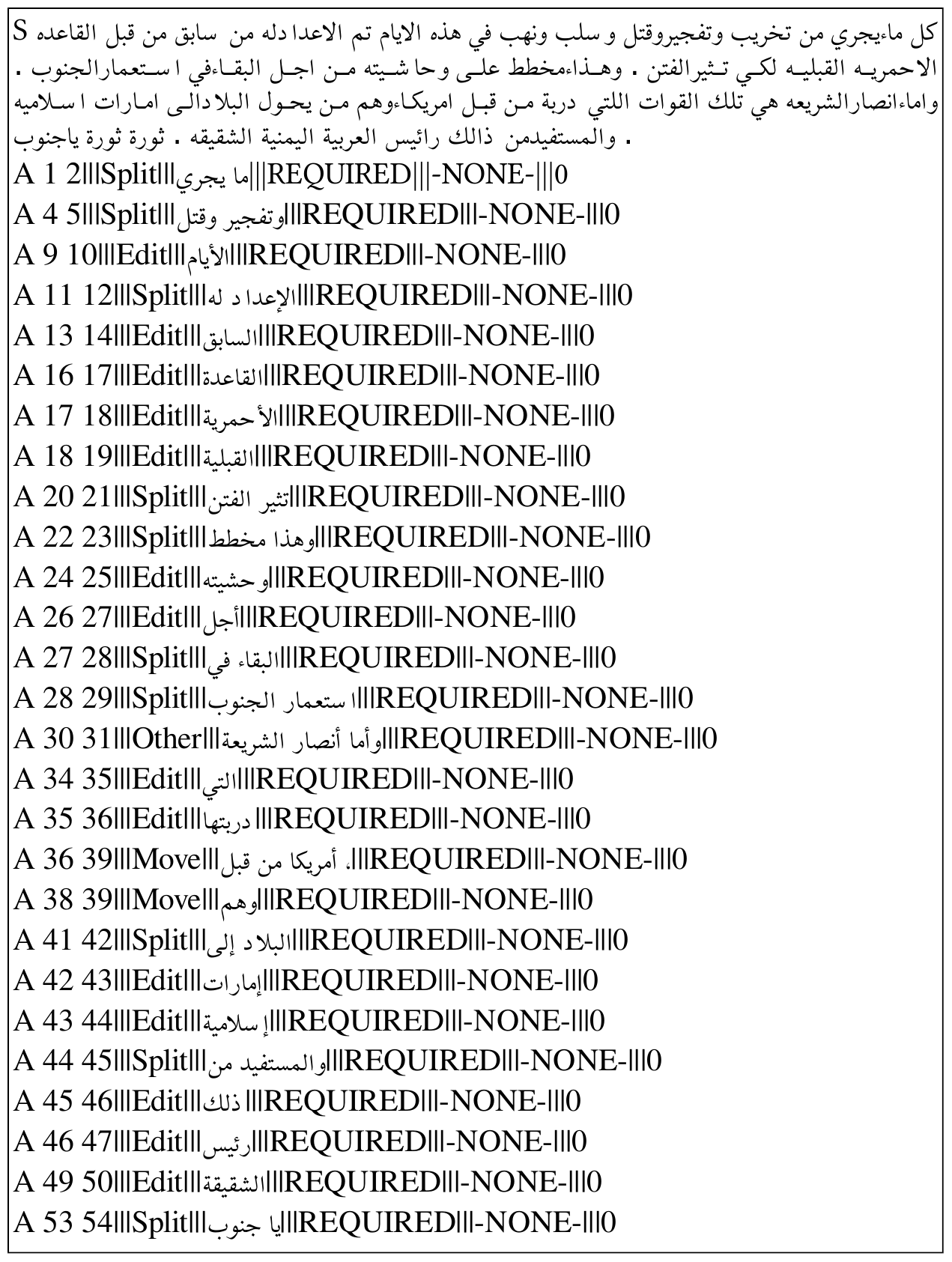}
\caption{An annotation example from the QALB corpus. Following line $S$ which refers to the potentially wrong original text, the correction actions are added for each incorrect token in $A$ lines.}
\label{app_qabl_sample}
\end{figure}

\newpage

\section{Error calculation}
\label{error_calculation}
\begin{lstlisting}[caption=Calculating error in training the RNN model]
def get_error(input_sequence, output_sequence):
		error = list()
    dy.renew_cg()
    embedded_sequence = embed_string(input_sequence)

    rnn_state = network.initial_state()
    rnn_outputs = rnn_run(rnn_state, embedded_sequence)

    for rnn_output, output_char in zip(rnn_outputs, output_sequence):
        w_out = dy.parameter(network.w_out)
        b_out = dy.parameter(network.b_out)
        probabilities = dy.softmax(output_w * rnn_output + output_b)
        error.append(-dy.log(dy.pick(probabilities, output_char)))
    error = dy.esum(error)
    return error
\end{lstlisting}

\begin{lstlisting}[caption=Calculating error in training the BRNN model]
def get_error(input_sequence, output_sequence):
    error = list()
    dy.renew_cg()
    embedded_sequence = embed_string(input_sequence)
    
    rnn_bwd_state = bwd_RNN.initial_state()
    rnn_bwd_outputs = run_rnn(rnn_bwd_state, embedded_sequence[::-1])[::-1]
    rnn_fwd_state = fwd_RNN.initial_state()
    rnn_fwd_outputs = run_rnn(rnn_fwd_state, embedded_sequence)
    rnn_outputs = [dy.concatenate([fwd_out, bwd_out]) for fwd_out, bwd_out in zip(rnn_fwd_outputs, rnn_bwd_outputs)]
    
    for rnn_output, output_char in zip(rnn_outputs, output_sequence):
        w_out = dy.parameter(network.w_out)
        b_out = dy.parameter(network.b_out)
        probabilities = dy.softmax(output_w * rnn_output + output_b)
        error.append(-dy.log(dy.pick(probabilities, output_char)))
	error = dy.esum(error)
	return error
\end{lstlisting}

\begin{lstlisting}[caption=Calculating error in training the encoder-decoder model]
def get_error(input_sequence, output_sequence):
    error = list()
    dy.renew_cg()
    embedded_sequence = embed_string(input_sequence)
    encoded_string = encode_string(embedded_sequence)[-1]
    rnn_state = decoder.initial_state()

    for output_char in output_sequence:
        rnn_state = rnn_state.add_input(encoded_string)
        w_out = dy.parameter(network.w_out)
        b_out = dy.parameter(network.b_out)
        probabilities = dy.softmax(output_w * rnn_state.output() + output_b)
        error.append(-dy.log(dy.pick(probabilities, output_char)))
    error = dy.esum(error)
    return error
\end{lstlisting}

\begin{lstlisting}[caption=Calculating error in training the attention-based encoder-decoder model]
def get_error(self, input_sequence, output_sequence):
    error = list()
    dy.renew_cg()
    embedded_sequence = embed_string(input_sequence)
    encoded_string = encode_string(embedded_sequence)
    rnn_state = decoder.initial_state().add_input(dy.vecInput(self.encoder_state_size))

    for output_char in output_sequence:
        attended_encoding = self._attend(encoded_string, rnn_state)
        rnn_state = rnn_state.add_input(attended_encoding)
        w_out = dy.parameter(network.w_out)
        b_out = dy.parameter(network.b_out)
        probabilities = dy.softmax(output_w * rnn_state.output() + output_b)
        error.append(-dy.log(dy.pick(probabilities, output_char)))
    error = dy.esum(error)
    return error
\end{lstlisting}


\newpage

\section{Qualitative comparison of the correction models}
\begin{figure}[!h]
\centering
\includegraphics[scale=0.9]{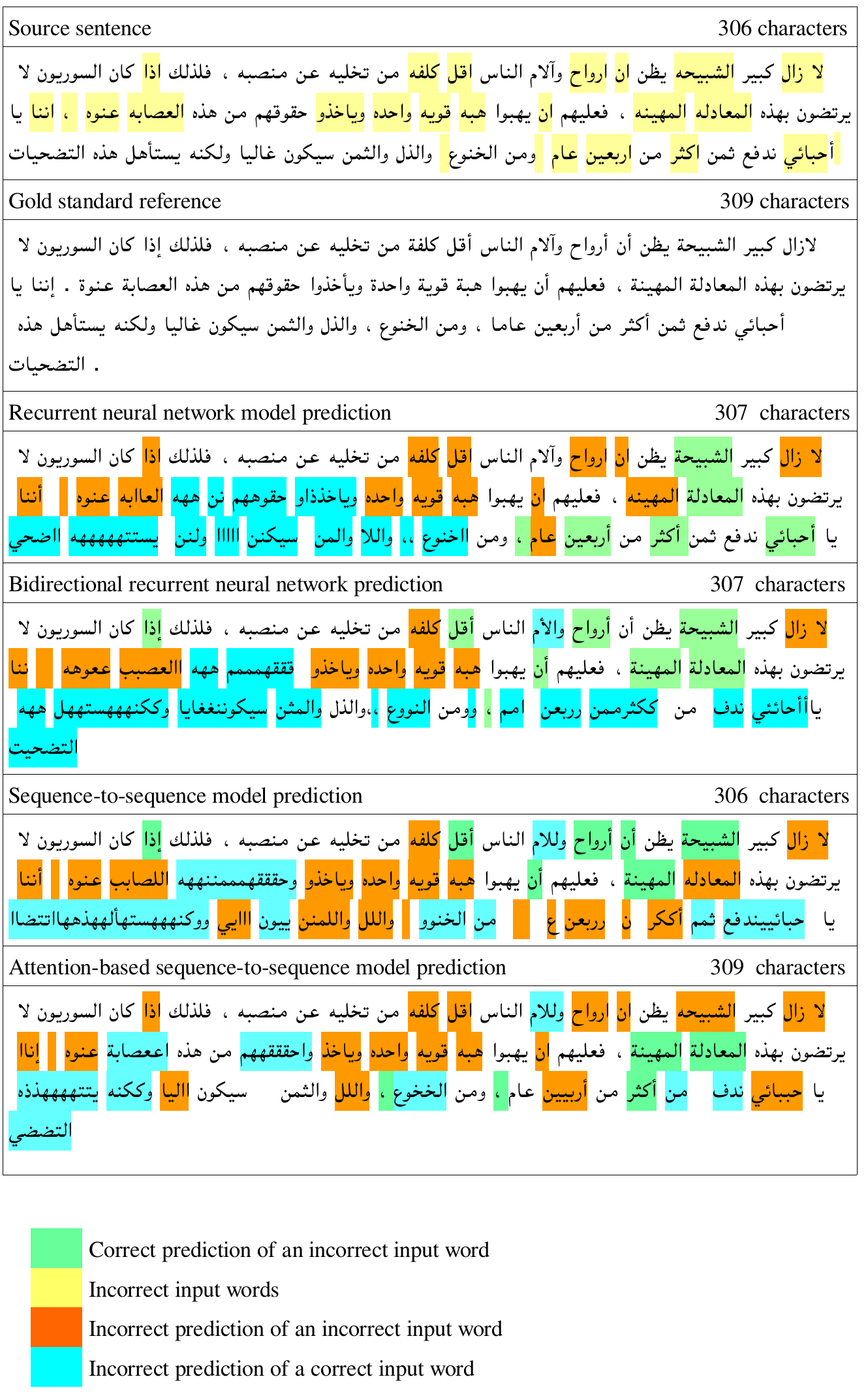}
\caption{Correction of an input using different models. The size of each sequence is marked in the header of each sequence.}
\label{app_qabl_sample}
\end{figure}

\label{reference}
\addcontentsline{toc}{chapter}{Bibliography}
\bibliography{references}{}
\bibliographystyle{unsrt}

\end{document}